\documentclass[sn-mathphys]{sn-jnl}

\jyear{2021}

\theoremstyle{thmstyleone}
\newtheorem{theorem}{Theorem}

\newtheorem{lemma}[theorem]{Lemma}

\theoremstyle{thmstyletwo}

\newtheorem{remark}{Remark}

\theoremstyle{thmstylethree}
\newtheorem{definition}{Definition}
\newtheorem{assumption}[definition]{Assumption}

\usepackage{benstyleSpringer}
\usepackage{enumerate}

\begin{document}

\title[Stable, accurate and efficient neural networks for inverse problems]{NESTANets: Stable, accurate and efficient neural networks for analysis-sparse inverse problems}

\author*[1]{\fnm{Maksym} \sur{Neyra-Nesterenko}}\email{mneyrane@sfu.ca}

\author[1]{\fnm{Ben} \sur{Adcock}}\email{ben\_adcock@sfu.ca}

\affil[1]{\orgdiv{Department of Mathematics}, \orgname{Simon Fraser University}, \country{Canada}}

\abstract{
Solving inverse problems is a fundamental component of science, engineering and mathematics.
With the advent of deep learning, deep neural networks have significant potential to outperform existing state-of-the-art, model-based methods for solving inverse problems.
However, it is known that current data-driven approaches face several key issues, notably hallucinations, instabilities and unpredictable generalization, with potential impact in critical tasks such as medical imaging. 
This raises the key question of whether or not one can construct deep neural networks for inverse problems with explicit stability and accuracy guarantees.
In this work, we present a novel construction of accurate, stable and efficient neural networks for inverse problems with general analysis-sparse models, termed NESTANets.
To construct the network, we first unroll NESTA, an accelerated first-order method for convex optimization. The slow convergence of this method leads to deep networks with low efficiency. Therefore, to obtain shallow, and consequently more efficient, networks we combine NESTA with a novel restart scheme. We then use compressed sensing techniques to demonstrate accuracy and stability. 
We showcase this approach in the case of Fourier imaging, and verify its stability and performance via a series of numerical experiments.
The key impact of this work is demonstrating the construction of efficient neural networks based on unrolling with guaranteed stability and accuracy.
}


\maketitle

\section{Introduction}\label{sec1}

Solving inverse problems is a crucial task in a wide range of scientific, industrial and medical applications. In this work, we focus on linear inverse problems taking the form:
\begin{align*}
\mbox{Given measurements $y = Ax + e$, recover $x$.}
\end{align*}
Here $A$ is a linear operator (known as the \textit{forward} or \textit{measurement} operator) and $e$ is a noise term. 
We consider the discrete version of this problem where $x \in \bbC^N$, the measurements and noise $y, e \in \bbC^m$ and the measurement matrix $A \in \bbC^{m \times N}$.

One of the key challenges in applications involving inverse problems is the highly underdetermined setting. Here $m$, the number of measurements, is substantially smaller than the ambient dimension $N$. In the case of image reconstruction, this is often termed \textit{compressive imaging} \cite{adcock2021compressive}. An effective way to tackle this problem involves exploiting some inherent low-dimensionality of the underlying object to be recovered. Over the last decade, regularization methods (often termed \textit{model-based} or \textit{handcrafted} methods) have become standard tools \cite{adcock2021compressive,arridge2019solving}. In such methods, one considers a fixed low-dimensional structure -- for example, sparsity or structured sparsity in an orthonormal wavelet basis, discrete gradient sparsity, or sparsity in a redundant dictionary -- and then designs a regularization term (often convex) to exploit such structure; for instance, the analysis or synthesis $\ell^1$-norm or the TV-norm, or some combination thereof. Such methods, whose performance is mathematically supported by the theory of \textit{compressed sensing}, have become standard in various applications, notably medical imaging \cite{adcock2021compressive,ravishankar2020image,FDAGE,FDASiemens}.

\subsection{Deep neural networks and deep learning for inverse problems}

Spurred by stunning successes in traditional machine learning tasks (e.g., classification), in the last five years there has been a significant focus on using deep learning for inverse problems \cite{wang2018image}. Typically, in this line of work a deep neural network reconstruction map $\cN : \bbC^m \rightarrow \bbC^N$ is learned from training data (pairs of images and their measurements). As opposed to model-based methods, which impose a fixed structure, deep learning methods implicitly learn it from the training data. 

Such methods have achieved some notable successes, with claimed performance gains over existing state-of-the-art model-based methods in a range of different settings. This field is rapidly evolving, therefore we will not attempt to survey it in full. See, e.g., \cite{adcock2021compressive,arridge2019solving,knoll2020deep,liang2020deep,lundervold2019overview,mccann2017convolutional,ongie2020deep,shen2020introduction,zhang2020review} and references therein for reviews of deep learning for inverse problems in imaging.

\subsection{Hallucinations, instabilities and unpredictable generalization}

However, there is growing concern that deep learning for inverse problems may suffer from several drawbacks, despite its seeming ability to achieve higher-quality reconstructions than state-of-the-art model-based methods. First, such methods can suffer from \textit{hallucinations}. A hallucination is the insertion of a false (i.e., one not present in the ground truth image) realistic feature in an image by a reconstruction map.
 Second, certain trained networks can be unstable. Hence, a small perturbation to the measurements $y$ (sometimes even a random perturbation \cite{gottschling2020troublesome}) can cause significant artefacts in the reconstructed image. Third, such trained networks can also suffer from unpredictable generalization behaviours outside of their training sets. For instance, simply increasing the number of measurements $m$ can lead to unexpectedly worse performance \cite{antun2020instabilities} -- an undesirable behaviour generally not seen in standard model-based methods.

These issues were first studied in \cite{huang2018some,antun2020instabilities} (see also \cite{gottschling2020troublesome} and \cite[Chpt.\ 20]{adcock2021compressive}). There is now growing concern over how these issues may prohibit the deployment of deep learning methods in practical imaging systems. In MRI (Magnetic Resonance Imaging), for instance, the issue of hallucinations has been discussed in the fastMRI challenge \cite{muckley2021results}, where they are referred to as `not acceptable' and `especially problematic'.\footnote{The full quote reads: ``Such hallucinatory features are not acceptable and especially problematic if they mimic normal structures that are either not present or actually abnormal.''} Similar sentiments have been expressed in the case of microscopy \cite{hoffman2021promise}, fluorescence microscopy \cite{belthangady2019applications}, PET (Position Emission Tomography) \cite{herskovits2021artificial}  and computed tomography \cite{bhadra2021hallucinations,sidky2021do}.\footnote{For example, in the work \textit{On hallucinations in tomographic image reconstruction} \cite{bhadra2021hallucinations} the authors write ``The potential lack of generalization of deep learning-based reconstruction methods as well as their innate unstable nature may cause false structures to appear in the reconstructed image that are absent in the object being imaged''.} 
 Suffice to say, these issues are currently the subject of active debate in the community. We refer to 
 \cite{boche2022limitations, hardy2021intraprocedural, herskovits2021artificial,  laine2021avoiding, larson2021regulatory, liu2022medical, morshuis2022adversarial, pal2022review, qi2021synergistic,  singh2020artificial,torres-velazquez2020application, varoquaux2022machine, yu2022validation,  wu2021medical, noordman2021,stumpo2022machine,tolle2021mean,lahiri2021learning,shimron2022implicit,gottschling2020troublesome} for further discussion on this topic and related issues.

\rem{
Following on from \cite{huang2018some,antun2020instabilities}, several recent works have also started to re-examine the stability of model-based reconstruction methods \cite{genzel2022solving,zhang2021instabilities, darestani2021measuring,alaifari2022localized}. Interestingly, these methods can also be quite susceptible to adversarial perturbations. In other words, some of the aforementioned phenomena are not exclusive to learning-based approaches (although it is arguable that learning may encourage such effects, e.g., hallucinations). Some recent analysis has suggested that this may be due to the moderate ill-posedness of the problem in certain applications, such as parallel MRI \cite{zhang2021instabilities}, rather than the (model-based or learning-based) reconstruction method employed. On the other hand, other works have suggested better robustness of model-based methods to worst-case noise \cite{colbrook2022difficulty},\cite[Chpt.\ 21]{adcock2021compressive}. Thus, there may be subtle effects due to the numerical implementation of the solvers that can lead to apparent changes in robustness.
Interestingly, \cite{genzel2022solving} shows an empirical \textit{accuracy-stability tradeoff} for deep learning: adding random noise to the training data increases the stability of the learned reconstruction map, but at the price of worse accuracy. This has also been described theoretically in \cite{gottschling2020troublesome}.  It is worth emphasizing at this stage that not all learning-based methods lead to unstable reconstruction schemes. Indeed, constructions such as \cite{hasannasab2020parseval,combettes2020lipschitz} lead to provably stable deep neural networks.
}

\subsection{Contributions and related work}

These issues lead to the following question, which is the starting point for this work: \textit{is it possible to construct deep neural networks for inverse problems with guarantees on stability and performance?} In particular, can we construct stable deep neural networks that perform at least as well as state-of-the-art model-based methods, thereby showing that they achieve the currently best known accuracy-stability tradeoff? If so, are these networks \textit{computationally efficient}? Specifically, are they not too wide or deep, so that forwards propagations (i.e., the reconstruction $y \mapsto \cN(y)$) can be performed fast? 

This question was first considered in \cite{colbrook2022difficulty} (see also \cite[Chpt.\ 21]{adcock2021compressive}). In \cite{colbrook2022difficulty}, the authors studied the case of Fourier imaging with the Discrete Fourier Transform (DFT) and binary imaging with the Discrete Walsh--Hadamard Transform (DHT). To provide performance guarantees, they considered a model class consisting of images that are approximately sparse in levels in the discrete {orthonormal} Haar wavelet basis \cite{adcock2016note}. Then, leveraging tools from compressed sensing theory and convex optimization, they showed that one can construct a neural network that recovered such images up to error depending on the distance to the model class (i.e., accuracy), the measurement noise (i.e., stability) and a term decaying exponentially fast in the number of network layers (i.e., efficiency). The precise construction of this network was achieved by \textit{unrolling} (see below) a certain weighted $\ell^1$-minimization problem using the primal-dual iteration (also known as the Chambolle--Pock algorithm) \cite{chambolle2011first,chambolle2016ergodic} in combination with a restart procedure. See, e.g., \cite{roulet2020sharpness,renegar2022simple} for further information on restarting techniques in optimization and \cite{colbrook2022warpd} for a restart scheme based on the primal-dual iteration. The latter extends \cite{colbrook2022difficulty} to general inverse problems satisfying approximate sharpness conditions.

In this paper, we extend and modify the work of \cite{colbrook2022difficulty} in the following ways. First, we consider model classes of images that are approximately analysis-sparse in a general redundant dictionary forming a frame. 
It is well-known that the {orthonormal} Haar wavelet transform is a relatively poor sparsifying transform for imaging.
Our setup includes many other orthogonal and nonorthogonal transforms. In particular, in our examples we consider the concatenated \textit{wavelets-plus-gradient} transform, which is commonly used in medical imaging applications \cite{fessler2020optimization,lustig2008compressed,lustig2007sparse}, including in commercially-available MR scanners \cite{fessler2020optimization}. Other common transforms, such as curvelets \cite{candes2000curvelets---a,candes2002recovering,candes2004tight}, ridgelets \cite{candes1999ridgelets:} or shearlets \cite{labate2005sparse,guo2006sparse,guo2007optimally,kutyniok2011compactly} could also be considered. Second, we unroll the NESTA algorithm, as opposed to the primal-dual iteration. NESTA is a well known algorithm for constrained $\ell^1$-minimization \cite{becker2011nesta,becker2011practical}. We unroll and combine it with a restart scheme based on \cite{roulet2020computational} that adjusts the NESTA smoothing parameter to achieve exponential convergence in the number of network layers. We refer to the resulting neural network construction as NESTANet. Note that our scheme extends that of \cite{roulet2020computational} from the exact sparsity and noiseless measurement setting, to the more general case of approximate (i.e., inexact) analysis sparsity and noisy measurements.

Like \cite{colbrook2022difficulty}, our work involves \textit{unrolling} an optimization algorithm. Unrolling is an important concept in deep learning for inverse problems, one which has been used in some of the best-performing networks. See, e.g., \cite{monga2021algorithm}, as well as \cite[Chpt.\ 21]{adcock2021compressive} and \cite{arridge2019solving,liang2020deep,mccann2017convolutional,ravishankar2020image}. For imaging, various different optimization algorithms have been unrolled. These include ISTA \cite{gregor2010learning,zhang2018ista}, ADMM \cite{yang2016deep,yang2020admm,arvinte2021deep}, the primal-dual iteration \cite{adler2018learned,wang2016proximal,colbrook2022difficulty}, block coordinate descent \cite{chun2018deep}, POCS \cite{cui2021equilibrated}, as well as approximate message passing algorithms \cite{metzler2017learned}, Neumann series \cite{gilton2020neumann}, field of experts models \cite{hammernik2018learning} and various others, including \cite{gilton2021deep}.  While successful, we remark that simply unrolling an optimization solver does not ensure stability and robustness of a deep learning procedure \cite{gottschling2020troublesome}. Also, to the best of our knowledge, unrolled NESTA has not previously been considered. 

It is important to stress that our work and \cite{colbrook2022difficulty} detail the construction of deep neural networks possessing accuracy and stability guarantees for certain model classes and inverse problems. In other words, they provide a quantification of the accuracy-stability tradeoff for certain problem classes. They do not claim that such networks can be learned from data. Therefore they do not directly explain the practical performance of deep learning -- a question which remains largely open. Nonetheless, they demonstrate that state-of-the-art performance is obtainable using architectures and principles (i.e., unrolling) that are similar to those used in practice.

\subsection{Outline}

In what follows, Section \ref{sec:main-res} sets up and states the main result of the paper.
In Section \ref{sec:restarted-NESTA}, we introduce restarted NESTA and prove recovery guarantees for the approximate analysis-sparse solution it computes.
Section \ref{sec:unrolling} presents the unrolled neural network construction of NESTANet, which leads into a proof of the main result in Section \ref{sec:final-args}. 
Section \ref{sec:experiments} covers numerical experiments, which showcase restarted NESTA in a practical setting.
The experiments validate theoretical results pertaining to accuracy, stability and efficiency, making it plausible to use in practice.
In addition, we offer guidelines for choosing hyperparameters and highlight practical differences between having and not having a restart scheme.
The final concluding section discusses prospective research.

\section{Results and discussion}\label{sec:main-res}

As mentioned, we consider the discrete linear inverse problem 
\be{
\label{inv-prob-main}
\mbox{Given measurements $y = Ax + e$, recover $x$.}
}
We assume that $A \in \bbC^{m \times N}$, $x \in \bbC^N$ and $y, e \in \bbC^m$. Furthermore, we now assume that the noise vector $e$ is bounded, and satisfies $\nm{e}_{\ell^2} \leq \eta$ for some positive constant $\eta$.

\subsection{Main result: stable, accurate and efficient deep neural networks for analysis-sparse models}\label{ss:main-res}

We now introduce some notation. We let $[M] = \{1,\ldots,M\}$ and, for $x = (x_i)^{M}_{i=1} \in \bbC^M$, we let $x_S$ denote the vector with $i$th entry equal to $x_i$ if $i \in S$ and zero otherwise.
We define the $\ell^1$-norm \textit{best $s$-term approximation} of $z \in \bbC^M$ by
$$ \sigma_s(z)_{\ell^1} = \min \left\{ \nmu{z - u_S}_{\ell^1} \ : \  u \in \bbC^M,\ S \subseteq [M] \right\},$$
where we note that $\sigma_s(z)_{\ell^1} = 0$ whenever $z$ is $s$-sparse.
We consider solutions to \R{inv-prob-main} that are approximately analysis-sparse with respect to an analysis operator $W \in \bbC^{N \times M}$.

Recall that the columns of $W$ form a frame for $\bbC^N$ if there exist constants $\beta \geq \alpha > 0$ such that
\bes{
\alpha \nm{x}^2_{\ell^2} \leq \nm{W^* x}^2_{\ell^2} \leq \beta \nm{x}^2_{\ell^2}.
}
In particular, we must have $M \geq N$.
Notice that the optimal values of $\alpha$ and $\beta$ are precisely $\alpha = (\sigma_{\min}(W))^2 = (\sigma_N(W))^2$ and $\beta = (\sigma_{\max}(W))^2 = (\sigma_{1}(W))^2 = \nm{W}^2_{\ell^2}$.
We assume that $\alpha = 1$ for convenience, noting that this can always be ensured by scaling the frame vectors.
With this, we make the following assumption regarding the operator $W$:
\begin{assumption}\label{ass:W-frame}
The columns of $W$ form a frame for $\bbC^N$ with lower frame bound $1$.
\end{assumption}

With this, we can now define the precise model class we seek to recover. Fix $\eta \geq 0$, $1 \leq s \leq M$ and define
\bes{
\mathcal{CS}_s(W^* x,\eta) = \frac{\sigma_s(W^* x)_{\ell^1}}{\sqrt{s}} + \eta.
}
Then, given $\chi > 0$ we let
$$
\bbI = \bbI_{W,\chi,\eta} = \left\{(x,e) \in \bbC^{N} \times \bbC^m : \nm{x}_{\ell^2} \leq 1,\ \nm{e}_{\ell^2} \leq \eta,\ \cC\cS_s(W^* x,\eta) \leq \chi \right\},
$$
and define the class of measurement vectors
\be{
\label{M-class}
\bbM = \bbM_{A,W,\chi,\eta} = \{y = A x + e \in \bbC^m : (x,e) \in \bbI_{W,\chi,\eta} \}.
}
In other words, we consider measurement vectors $y = A x + e$ corresponding to noisy measurements of signals $x \in \bbC^N$ that are approximately sparse (in the sense that $\sigma_{s}(W^* x)_{\ell^1} / \sqrt{s} \leq \chi$) in the given analysis model. With this in hand, our aim is to demonstrate the existence of a deep neural network that recovers any such $x$ up to the parameter $\chi$, i.e.\ accurately (due to the $\sigma_{s}(W^* x)_{\ell^1} / \sqrt{s}$ term) and stably (due to the $\eta$ term).

In order to do this, we also need conditions of the measurement matrix $A$ and analysis operator $W$. We consider the following standard condition:

\defn{
The matrices $A \in \bbC^{m \times N}$ and $W \in \bbC^{N \times M}$ satisfy the \textit{robust Null Space Property (rNSP)} with constants $0 < \rho < 1$ and $\gamma > 0$ if
\bes{
\nmu{(W^* v)_S}_{\ell^2} \leq \frac{\rho}{\sqrt{s}} \nm{(W^* v)_{S^c}}_{\ell^1} + \gamma \nm{A v}_{\ell^2},
} 
for all $v \in \bbC^N$ and $S \subseteq [M]$ with $\vert S \vert \leq s$.
}

See, e.g., \cite[Defn.\ 3]{foucart2014stability} (note that this definition corresponds to case $p = 2$ and $\nm{\cdot} = \nm{\cdot}_{\ell^2}$ considered therein).

In addition, we impose the condition that $A$ has orthonormal rows up to a constant factor, i.e. $AA^* = \nu I$ for some $\nu > 0$. In the discussion that follows from stating the main result, we elaborate more on this requirement.
We now state our main result:

\thm{
[Stable, accurate and efficient deep neural networks]
\label{thm:main-res}
Let $W \in \bbC^{N \times M}$ satisfy Assumption \ref{ass:W-frame} with upper frame constant $\beta \geq 1$ and $A \in \bbC^{m \times N}$ with $m \leq N$ be such that the pair $(A,W)$ has the rNSP of order $1 \leq s \leq M$ with constants $0 < \rho < 1$ and $\gamma > 0$. Suppose also that $A A^* = \nu I$ for some $\nu > 0$. Let $\eta \geq 0 $, $\chi > 0$ and consider the class $\bbM = \bbM_{A,W,\chi,\eta}$ defined in \R{M-class}. Then, for every $0 < r < 1$ and $k \geq 1$ one can construct a deep neural network (NESTANet) $\cN : \bbC^m \rightarrow \bbC^N$ such that
\bes{
\nmu{x - \cN(y)}_{\ell^2} \leq c_1 \cdot \chi + r^{k},\quad \forall y = Ax+e \in \bbM,
}
where $c_1$ depends on $\rho$, $\gamma$, and $r$ only. The number of layers in the network is bounded by $c_2 \cdot \sqrt{M/s} \cdot k$, where $c_2$ depends on $\rho$, $\gamma$, $\beta$ and $r$ only, and its width is bounded by $3N+M$. 
}

For ease of presentation, we do not describe the precise architecture of the deep neural networks asserted in this theorem. See Section \ref{ss:DNN-class} and Theorem \ref{thm:unroll-restarted-NESTA} for the detailed construction.

\subsection{Discussion}

There are several key points worth mentioning about Theorem \ref{thm:main-res}, specifically how it can be generalized and comparisons to the influential work in \cite{colbrook2022difficulty}.

First, the error bound presented in Theorem \ref{thm:main-res} tells us that the best achievable error bound will be proportional to $\chi$.
Choosing $k = \lceil \log(\chi) / \log(r) \rceil$ yields a network that reconstructs $y \in \bbM$ to its associated vector $x$ within an error proportional to the desired error.
This is a measure of efficiency for our network construction, where in order to guarantee reconstruction within an error proportional to $\chi$, the required depth of the network scales logarithmically in $\chi$. 
In particular, this is analogous to what is observed in \cite[Thm. 5.10]{colbrook2022difficulty}.
Further drawing a comparison to \cite[Thm. 5.5]{colbrook2022difficulty}, their network construction has depth proportional to $np$ layers where $n$ is the iteration number and $p \propto \nm{A}_{\ell^2}$. It may be tempting to think that their construction involves substantially fewer layers than that of Theorem \ref{thm:main-res}, where the number of layers depends on $\sqrt{N/s}$ when $N = M$ (the case studied in \cite{colbrook2022difficulty}). This is typically not the case, since $\nm{A}_{\ell^2}$ has an upper bound proportional $\sqrt{N/s}$ whenever $A$ has the \textit{restricted isometry property} (see \cite[Rem. 8.8]{adcock2021compressive}), a stronger condition implying the rNSP.

Second, the condition $A A^* = \nu I$ is required for the NESTA algorithm, and is thus needed when constructing NESTANets by unrolling NESTA. Note that this condition can be relaxed within the NESTA algorithm if the singular value decomposition of $A$ can be obtained \cite[Sec.\ 3.7]{becker2011practical}. However, taking this approach additionally involves solving a nonlinear equation in every iteration of NESTA, so it is unclear how this can be unrolled into a neural network.
Although seemingly restrictive, this condition on $A$ is common in compressed sensing and arises when constructing certain matrices $A$ to satisfy the rNSP, e.g., subsampled unitary matrices (see, e.g., \cite[Sec.\ 5.3.2]{adcock2021compressive}). Such matrices arise frequently in applications, particularly in imaging.

Third, we note that Theorem \ref{thm:main-res} only considers the sparse model. It is also possible to consider certain structured sparsity models -- for example, the \textit{sparsity in levels} model \cite[Chpt.\ 12]{adcock2021compressive} that was considered in \cite{colbrook2022difficulty}. In the setting of Theorem \ref{thm:main-res}, doing so would simply require replacing the rNSP assumption to a suitable structured rNSP.

\section{Restarted NESTA algorithm}\label{sec:restarted-NESTA}

The remainder of this paper is devoted to the construction of NESTANet, the deep neural network that yields the proof of Theorem \ref{thm:main-res}. As noted, this is based on unrolling a restarted variant of the NESTA algorithm for solving a certain analysis $\ell^1$-minimization problem. In this section, we describe this problem and algorithm. Then, in Section \ref{sec:unrolling} we show how this can be formulated as a deep neural network. Finally, we conclude in Section \ref{sec:final-args} with the proof of Theorem \ref{thm:main-res}.

\begin{algorithm}[t] \caption{NESTA for \R{eqn:W-QCBP}}\label{alg:NESTA} 
\begin{algorithmic}[1]
\Require Measurements $y \in \bbC^m$, measurement matrix $A \in \bbC^{m \times N}$ with $AA^* = \nu I$, analysis matrix $W \in \bbC^{N \times M}$, parameter $\beta \geq \nm{W}^2_{\ell^2}$,
 parameter $\eta > 0$, smoothing parameter $\mu$, maximum number of iterations $n_{\max}$, sequences $\{\alpha_n\}_{n=0}^{n_{\max}}$, $\{\tau_n\}_{n=0}^{n_{\max}}$, initial estimate $z_0$
\Ensure The vector $x_{n_{\max}}$, an approximate minimizer of \R{eqn:W-QCBP}
  \For{$n = 0, 1, \dots, n_{\max}$}
\State Compute $x_n$:
\State $q \leftarrow z_n - \frac{\mu}{\beta} W \cT_{\mu}(W^* z_n)$
\State $\lambda \leftarrow \max \{ 0, \eta^{-1} \nm{y - Aq}_{\ell^2} - 1 \}$
\State $x_n \leftarrow \left(I - \frac{\lambda}{(\lambda + 1)c} A^* A \right) \left(\frac{\lambda}{c} A^* y + q \right)$
\State Compute $v_n$: 
\State $q \leftarrow z_0 - \frac{\mu}{\beta} \sum_{i=0}^{n} \alpha_i W \cT_{\mu}(W^* z_n)$
\State $\lambda \leftarrow \max \{ 0, \eta^{-1} \nm{y - Aq}_{\ell^2} - 1 \}$
\State $v_n \leftarrow \left(I - \frac{\lambda}{(\lambda + 1)c} A^* A \right) \left(\frac{\lambda}{c} A^* y + q \right)$
\State Compute $z_{n+1} \leftarrow \tau_n v_n + (1 - \tau_n) x_n$
 \EndFor
\end{algorithmic}
\end{algorithm}

NESTA is a well-known algorithm for solving the constrained analysis $\ell^1$-minimization problem
\be{
  \label{eqn:W-QCBP}
    \min_{x \in \bbC^N} \nm{W^*x}_{\ell^1} \quad \mbox{subject to $\nm{y - Ax}_{\ell^2} \leq \eta$},
}
which is often referred to as \textit{analysis Quadratically Constrained Basis Pursuit (QCBP).} Introduced in \cite{becker2011nesta} (see also \cite{becker2011practical}), it is based on a combination of Nesterov's accelerated projected gradient descent algorithm \cite{nesterov1983method}, also known as Nesterov's method, and smoothing \cite{nesterov2005smooth}. Specifically, in \cite{becker2011nesta} the problem \R{eqn:W-QCBP} is first replaced by the smoothed problem 
\begin{align}
    \min_{x \in \bbC^N} \nm{W^*x}_{\ell^1,\mu} \quad \mbox{subject to $\nm{y - Ax}_{\ell^2} \leq \eta$}, \label{eqn:W-QCBP-smoothed}
\end{align}
where $\mu > 0$ is a smoothing parameter. Here
\bes{
\nm{x}_{\ell^1,\mu} = \sum^{M}_{i=1} H_{\mu}(x_i) ,\qquad H_{\mu}(x) : = \begin{cases} \frac{1}{2 \mu} \abs{x}^2 & \abs{x} \leq \mu \\ \abs{x} - \frac{\mu}{2} & \abs{x}  > \mu \end{cases} ,
}
is the smoothed $\ell^1$-norm (the Moreau envelope of the $\ell^1$-norm) and $H_{\mu}$ is the (complex) Huber function. Next, Nesterov's projected gradient descent algorithm \cite{nesterov1983method} is applied to \R{eqn:W-QCBP}. Let 
$\cT_{\mu} : \bbC^M \rightarrow \bbC^M$ be defined by
\bes{
(\cT_{\mu}(x_i)) = \begin{cases} x_i / \mu & \abs{x_i} \leq \mu \\ \sgn(x_i) & \abs{x_i} > \mu \end{cases},\quad x = (x_i)^{M}_{i=1} \in \bbC^M
}
(in the real case, this is simply the gradient of $H_{\mu}$). Using this in combination with an explicit formula for the projection onto the feasible set of \R{eqn:W-QCBP} (which is available since $A A^* = \nu I$) leads to NESTA. We omit the details of its derivation (see, e.g., \cite{becker2011nesta,becker2011practical} or \cite[Sec.\ 7.6]{adcock2021compressive}), and simply summarize it as Algorithm \ref{alg:NESTA}. Note that the original presentations of NESTA and Nesterov's method are done with real numbers only. To derive Algorithm \ref{alg:NESTA}, one represents a complex-valued vector as a real-valued vector, e.g. by concatenating the real and imaginary part vectors. After carefully deriving the algorithm over real-valued expressions, one reverts back to the equivalent complex-valued expressions.

\subsection{NESTA error bound}

We now give error bounds for the iterates produced by Algorithm \ref{alg:NESTA}. We commence with a bound for the objective function error:

\lem{ 
[Error bound for the objective function]
\label{lem:NESTA-W-QCBP-bound}
Let $x_n$ be the result of the $n$th iteration of Algorithm \ref{alg:NESTA} with initial vector $z_0$ and parameters $\alpha_i = \frac{i+1}{2}$ and $\tau_i = \frac{2}{i+3}$. Then
\bes{
\nmu{W^* x_n}_{\ell^1} - \nmu{W^* x}_{\ell^1} \leq \frac{2 \beta }{\mu (n+1)^2} \nmu{x - z_0}^2_{\ell^2} + \frac{M \mu}{2},\qquad \forall x : \nm{y - A x}_{\ell^2} \leq \eta.
}
}
\begin{proof}
To be precise, Algorithm \ref{alg:NESTA} is Nesterov's method applied to \eqref{eqn:W-QCBP-smoothed} with proximity function $d(x) = \frac{1}{2} \nmu{x - z_0}_{\ell^2}^2$, convex set $Q = \{x \in \bbC^{N} : \nmu{Ax - y}_{\ell^2} \leq \eta \}$, and objective function $f(x) = \nmu{W^* x}_{\ell^1,\mu}$.
The gradient of $f$ has Lipschitz constant $L = \beta / \mu$, since $W$ satisfies Assumption \ref{ass:W-frame}.
Then, by slightly modifying the proof of \cite[Thm.\ 2]{nesterov2005smooth}\footnote{Specifically, in \cite[Thm.\ 2]{nesterov2005smooth} the objective error bound is given in terms of $\hat{x}_{\mu}$, where $\hat{x}_{\mu}$ is an optimal solution of the smoothed problem \R{eqn:W-QCBP-smoothed}. But the proof does not use the optimality property in any way, only feasibility. It therefore applies to any feasible vector $x$. This minor modification is convenient for later developments.} we deduce that
\begin{align}
    \nmu{W^* x_n}_{\ell^1,\mu} - \nmu{W^*x}_{\ell^1,\mu}
    \leq \frac{2 \beta}{\mu (n+1)^2} \nmu{x - z_0}_{\ell^2}^2,\quad \forall x : \nm{y - A x}_{\ell^2} \leq \eta.
    \label{eqn:Nesterov bound for smooth W-QCBP}
\end{align}
Now, the Huber function satisfies
\begin{align*}
    H_{\mu}(a) \leq \avu{a} \leq H_{\mu}(a) + \frac{\mu}{2}, \qquad \forall a \in \bbC,
\end{align*}
thus for $\nm{z}_{\ell^1,\mu} = \sum_{i=1}^{M} H_{\mu}(z_i)$ we get
\begin{align}
    \nm{z}_{\ell^1,\mu} \leq \nm{z}_{\ell^1} \leq \nm{z}_{\ell^1,\mu} + \frac{M \mu}{2}, \qquad \forall z \in \bbC^{M}. \label{eqn:Huber inequality}
\end{align}
Combining this observation and \eqref{eqn:Nesterov bound for smooth W-QCBP} gives
\begin{align*}
    \nmu{W^* x_n}_{\ell^1} & \leq \nmu{W^* x_n}_{\ell^1,\mu} + \frac{M \mu}{2} 
    \\
    & \leq \nmu{W^* x}_{\ell^1,\mu} + \frac{2 \beta}{\mu (n+1)^2} \nmu{x - z_0}_{\ell^2}^2 + \frac{M \mu}{2}
    \\
    & \leq \nmu{W^* x}_{\ell^1} + \frac{2 \beta}{\mu (n+1)^2} \nmu{x - z_0}_{\ell^2}^2 + \frac{M \mu}{2} ,
\end{align*}
as required.
\end{proof}

\thm{
[Error bound for the iterates]
\label{thm:NESTA-rNSP-error}
Let $\eta > 0$, $A \in \bbC^{m \times N}$ satisfy $A A^* = \nu I$ for some $\nu > 0$ and $W \in \bbC^{N \times M}$ satisfy Assumption \ref{ass:W-frame}. Suppose that $A$ and $W$ satisfy the rNSP of order $1 \leq s \leq M$ with constants $0 < \rho < 1$ and $\gamma > 0$. Let $x \in \bbC^N$, $y = A x + e \in \bbC^m$, $\nm{e}_{\ell^2} \leq \eta$, and $x_n$ be the result of the $n$th iteration of Algorithm \ref{alg:NESTA} with initial vector $z_0$ and parameters $\alpha_i = \frac{i+1}{2}$ and $\tau_i = \frac{2}{i+3}$. Then
\bes{
\nm{x - x_n}_{\ell^2} \leq \nm{W^*(x - x_n)}_{\ell^2} \leq \zeta + c_1 \frac{M \mu}{2 \sqrt{s}} +  \frac{2 c_1 \beta}{\mu (n+1)^2 \sqrt{s}} \nm{x - z_0}^2_{\ell^2} ,
}
where $c_1 = \frac{(1+\rho)^2}{1-\rho}$, $c_2 = \frac{(3 + \rho) \gamma}{1-\rho}$ and
\bes{
\zeta = 2 c_1 \frac{\sigma_{s}(W^* x)_{\ell^1}}{\sqrt{s}} + 2 c_2 \eta.
}
}
\begin{proof}
Using, for instance, \cite[Lem.\ 6.24]{adcock2022sparse}, we see that 
\bes{
    \nmu{W^*(x - x_n)}_{\ell^2} \leq \frac{c_1}{\sqrt{s}} \left( \nmu{W^* x_n}_{\ell^1} - \nmu{W^* x}_{\ell^1} + 2 \sigma_s (W^* x)_{\ell^1} \right) + c_2 \nmu{A(x - x_n)}_{\ell^2}.
}
Since $x$ is feasible for \eqref{eqn:W-QCBP-smoothed} and the iterate $x_n$ produced Algorithm \ref{alg:NESTA} is also feasible, we have
\bes{
\nmu{A(x - x_n)}_{\ell^2} \leq \nm{A x - y}_{\ell^2} + \nm{A x_n - y}_{\ell^2} \leq 2 \eta.
}
Using the fact that $x$ is feasible once more, we next apply Lemma \ref{lem:NESTA-W-QCBP-bound} to get
\bes{
\nmu{W^* x_n}_{\ell^1} - \nmu{W^* x}_{\ell^1} \leq \frac{2 \beta}{\mu (n+1)^2} \nmu{x - z_0}^2_{\ell^2} + \frac{M \mu}{2} .
}
Substituting these two estimates into the previous expression now gives the result.
\end{proof}

\subsection{Restarted NESTA for \R{eqn:W-QCBP}}

\begin{algorithm}[t] \caption{Restarted NESTA for \R{eqn:W-QCBP}}\label{alg:restarted-NESTA} 
\begin{algorithmic}[1]
\Require Initial estimate $x^{\star}_0$, number of restarts $K$, sequences $\{\mu_k\}_{k=1}^{K+1}$, $\{n_k\}_{k=1}^{K+1}$.
\Ensure The vector $x^{\star}_{K+1}$, an approximate minimizer of \R{eqn:W-QCBP}
  \For{$k = 1, \dots, K+1$}
\State Set $x^{\star}_k$ as the output of NESTA (Algorithm \ref{alg:NESTA}) with initial estimate $x^{\star}_{k-1}$, smoothing parameter $\mu_k$ and number of iterations $n_k$.
 \EndFor
\end{algorithmic}
\end{algorithm}

A restart scheme is an algorithmic framework that can potentially accelerate convergence of an optimization algorithm.
The scheme has been applied to and analyzed for Nesterov's method in a general setting \cite{roulet2020sharpness}, where under suitable conditions one guarantees exponential convergence with respect to the restarts.

The idea behind a restart scheme is to notice that the error in Theorem \ref{thm:NESTA-rNSP-error} depends on the accuracy of the initial estimate $\nmu{x - z_0}_{\ell^2}$, the compressed sensing error $\zeta$ and the smoothing parameter $\mu$. In restarted NESTA, the algorithm is first run for a finite number of iterations with a fixed value of $\mu_1$ and some initial value. Then, the output is fed into the algorithm as a new initial value and the parameter $\mu_2$ appropriately scaled. This process is repeated for some number of iterations, e.g. until the objective function stops decreasing. We summarize restarted NESTA for \R{eqn:W-QCBP} in Algorithm \ref{alg:restarted-NESTA}. 

The main issue is how to choose the smoothing parameters $\mu_k$ and number of NESTA iterations $n_k$ to accelerate convergence. These are informed by the following result, which is based on Theorem \ref{thm:NESTA-rNSP-error}.

\thm{ \label{thm:NESTA-restart-error}
Let $A$, $W$, $x$, $y$, $\eta$, $s$, $c_1$, and $\zeta$ be as in Theorem \ref{thm:NESTA-rNSP-error}. Let $0 <  r < 1$ and define 
\bes{
\mu_k = \frac{r \sqrt{s}}{c_1 M}  \epsilon_{k-1},\quad n_k = \left \lceil  \frac{2 c_1 \sqrt{\beta M}}{r \sqrt{s}} \right \rceil - 1,
}
where $\epsilon_k$ is the sequence defined recursively as
\bes{
\epsilon_0 = \max \{ \nm{x}_{\ell^2} , \zeta \} ,\quad \epsilon_{k+1} = r \epsilon_k + \zeta.
}
Consider Algorithm \ref{alg:restarted-NESTA} with these values of $\mu_k$ and $n_k$ and initial value $x^{\star}_{0} = 0$. Then
\bes{
\nmu{x - x^{\star}_k }_{\ell^2} \leq \epsilon_k, \quad \epsilon_{k+1} = r^{k+1} \epsilon_0  + \frac{1-r^{k+1}}{1-r} \zeta,\quad k = 0,1,2,\ldots.
}
}
\prf{
We use induction on $k$. By construction, the result holds for $k = 0$. Now suppose the result holds for $k$. Then by Theorem \ref{thm:NESTA-rNSP-error},
\be{
\label{christmas-tree}
\nmu{x - x^{\star}_{k+1} }_{\ell^2} \leq \zeta + c_1 \frac{M \mu_{k+1}}{2 \sqrt{s}} + \frac{2 c_1 \beta}{\mu_{k+1} (n_{k+1}+1)^2 \sqrt{s}} \epsilon^2_k .
}
By definition of $\mu_{k+1}$ and $n_{k+1}$, we have
\bes{
c_1 \frac{M \mu_{k+1}}{2 \sqrt{s}} = \frac{r}{2} \epsilon_k,
}
and
\bes{
\frac{2 c_1 \beta}{\mu_{k+1} (n_{k+1}+1)^2 \sqrt{s} } \epsilon^2_{k} = \frac{2 c^2_1 M \beta}{(n_{k+1}+1)^2 r s } \epsilon_k \leq \frac{r}{2} \epsilon_k
}
Hence \R{christmas-tree} gives
\bes{
\nmu{x - x^{\star}_{k+1} }_{\ell^2} \leq \zeta + r \epsilon_k = \epsilon_{k+1},
}
as required.
}

\section{Unrolling}\label{sec:unrolling}

In this section, we describe the unrolling of (restarted) NESTA.

\subsection{The class of neural networks}\label{ss:DNN-class}

First, following \cite[Sec.\ 21.3.2]{adcock2021compressive} we define a general class of neural networks that is suitable for describing unrolled optimization schemes such as NESTA. We consider complex-valued feedforward neural networks $\cN : \bbC^m \rightarrow \bbC^N$ of the form
\begin{align*}
    \cN(y) = A^{(L)} \circ \sigma^{(L-1)} \circ A^{(L-1)} \circ \cdots \circ \sigma^{(1)} \circ A^{(1)}(y),
\end{align*}
where $L \geq 2$, and for each $l = 1, \dots, L$, $A^{(l)} : \bbC^{n_{l-1}} \rightarrow \bbC^{n_{l}}$ is an affine map of the form
\begin{align*}
    A^{(l)}(x) = W^{(l)}x + b^{(l)}(y), \qquad W^{(l)} \in \bbC^{n_l \times n_{l-1}},
\end{align*}
where $n_0 = m$ and $n_L = N$.
Here we allow the biases $b^{(l)}(y)$ to be affine maps of the input $y$, i.e.
\begin{align*}
    b^{(l)}(y) = R^{(l)}y + c^{(l)}, \qquad R^{(l)} \in \bbC^{n_l \times m}, \ c^{(l)} \in \bbC^{n_l}.
\end{align*}
While this is a slight departure from the standard feedforward setup, it is useful and standard when unrolling optimization schemes. See, e.g., \cite[Sec.\ 21.3.1]{adcock2021compressive}.
We also allow the activation functions $\sigma^{(l)} : \bbC^{n_l} \rightarrow \bbC^{n_l}$ to take one of the two following forms:
\begin{enumerate}[(i):]
\item  There is an index set $I^{(l)} \subseteq \{1, \dots, n_l\}$ such that $\sigma^{(l)}$ acts componentwise on those components of the input vector with indices in $I^{(l)}$ while leaving the rest unchanged.
\item  There is a nonlinear function $\rho^{(l)} : \bbC \rightarrow \bbC$ such that, if the input vector $x$ to the layer takes the form $x = (x_1, u, v)$, where $x_1$ is a scalar and $u$, $v$ are (possibly) vectors, then $\sigma^{(l)}(x) = (0, \rho^{(l)}(x_1) u, v)$.
\end{enumerate}
This is again useful when unrolling optimization schemes. For discussion on this assumption, see \cite[Sec.\ 21.3.2]{adcock2021compressive} and \cite{colbrook2022difficulty}. With this in hand, we denote the class of networks of this form as $\mathcal{N}_{\bm{n}, L, q}^*$, where
\begin{align*}
    \bm{n} = (n_0 = m, n_1, \dots, n_{L-1}, n_L = N),
\end{align*}
$L$ denotes the number of layers and $q$ is the number of different nonlinear activation functions used.

\subsection{Unrolled NESTA and NESTANets}

To proceed with network construction, we first construct simpler networks that compute the update steps of Algorithm \ref{alg:NESTA}. Afterwards, we combine the results to construct a network that computes one NESTA iteration, and from this we construct NESTANet, a network for Algorithm \ref{alg:restarted-NESTA}.
\begin{lemma} \label{lem:unroll-compute-q} 
Let $z_1, z_2 \in \bbC^N$ and $\alpha \in \bbC$. Then the map $\bbC^{2N} \rightarrow \bbC^N$ defined by
\begin{align*}
    \begin{pmatrix} z_1 \\ z_2 \end{pmatrix} 
    \mapsto z_1 - \alpha W \cT_\mu(W^* z_2)
\end{align*}
can be expressed as a neural network $\cN \in \cN_{\bm{n}, 2, 1}^*$ with $\bm{n} = (2N, N+M, N)$ and with all biases equal to zero, i.e. independent of the input.
\end{lemma}
\begin{proof}
Write this map as the following sequence of maps
\begin{align*}
    \begin{pmatrix} z_1 \\ z_2 \end{pmatrix}
    \overset{\mathrm{(a)}}{\longmapsto}
    \begin{pmatrix} z_1 \\ W^* z_2 \end{pmatrix}
    \overset{\mathrm{(b)}}{\longmapsto}
    \begin{pmatrix} z_1 \\ \cT_\mu(W^* z_2) \end{pmatrix}
    \overset{\mathrm{(c)}}{\longmapsto}
    z_1 - \alpha W \cT_\mu(W^* z_2).
\end{align*}
Here (a) and (c) are linear maps, noting that $W \in \bbC^{N \times M}$ is a fixed matrix.
The map (b) applies the complex Huber function with fixed parameter $\mu$, componentwise to the $M$ entries of $W^* z_2$.
Such a map corresponds to a nonlinear activation function of type (i). 
This gives the result.
\end{proof}

\begin{lemma}\label{lem:unroll-compute-lambda}
Fixing $y \in \bbC^m$, the map $\bbC^N \rightarrow \bbC$ defined by
\begin{align*}
    z \mapsto \max\{0,\eta^{-1}\nm{y - Az}_{\ell_2} - 1 \}
\end{align*}
can be expressed as a neural network $\cN \in \cN_{\bm{n}, 3, 2}^*$ with $\bm{n} = (N, m, 1, 1)$ and biases depending affinely on the vector $y$, but otherwise independent of the input.
\end{lemma}
\begin{proof}
For brevity, let $\sigma_1$ denote the squaring activation function $x \mapsto x^2$ and $\sigma_2$ denote the nonlinear activation function $x \mapsto \max\{0,\eta^{-1}\sqrt{x}-1\}$. 
Then we can express the map in question as the following sequence
\begin{align*}
    z 
    \overset{\mathrm{(a)}}{\longmapsto}
    y-Az
    \overset{\mathrm{(b)}}{\longmapsto}
    \sigma_1(y-Az)
    \overset{\mathrm{(c)}}{\longmapsto}
    \sigma_1(y-Az) \cdot \bm{1}
    \overset{\mathrm{(d)}}{\longmapsto}
    \sigma_2(\sigma_1(y-Az) \cdot \bm{1}),
\end{align*}
noting that $\sqrt{\sigma_1(y-Az) \cdot \bm{1}} = \nm{y-Az}_{\ell_2}$, where $\bm{1}$ is a vector of ones (of compatible dimension). 
Here (a) is an affine map and (c) is a linear map. 
The maps (b) and (d) apply the nonlinear activation functions $\sigma_1$ and $\sigma_2$, respectively.
Both activation functions are of type (i).
To ensure this sequence of maps corresponds to a network in $\cN_{\bm{n},3,2}^*$, the last map must be affine. We do this by appending the identity map at the end of the sequence.
Combining these observations gives the desired neural network.
\end{proof}

\begin{lemma}\label{lem:unroll-compute-x-v}
Let $u \in \bbC$, $w \in \bbC^N$, $y \in \bbC^m$ and $A \in \bbC^{m \times N}$ where $AA^* = \nu I$ for some $\nu > 0$.
Then the map $\bbC^{N+1} \rightarrow \bbC^N$ described by
\begin{align*}
    \begin{pmatrix} u \\ w \end{pmatrix} \mapsto \left(I - \frac{u}{(u+1)c} A^* A \right) \left(\frac{u}{c}A^* y + w \right)
\end{align*}
can be expressed as a neural network $\cN \in \cN_{\bm{n},2,1}^*$ with $\bm{n} = (N+1, 2N+1, N)$ and biases independent of the input. 
\end{lemma}
\begin{proof}
Observe that expanding the matrix product and using that $A A^* = c I$ yields
\begin{align*}
\left(I - \frac{u}{(u+1)c} A^* A \right) \left(\frac{u}{c}A^* b + w \right) = \frac{u}{(u+1)c} A^* (b - Aw) + w
\end{align*}
Let $\rho(u) = \frac{u}{u+1}$.
Proceeding, we have the sequence
\begin{align*}
    \begin{pmatrix} u \\ w \end{pmatrix}
    &
    \overset{\mathrm{(a)}}{\longmapsto}
    \begin{pmatrix} u \\ \frac{1}{c} A^*(b - A w) \\ w \end{pmatrix}
    \overset{\mathrm{(b)}}{\longmapsto}
    \begin{pmatrix} 0 \\ \frac{\rho(u)}{c} A^* (b - A w) \\ w \end{pmatrix}
    \overset{\mathrm{(c)}}{\longmapsto}
    \frac{\rho(u)}{c} A^* (b - A w) + w.
\end{align*}
The map (c) is linear, (a) is affine, and (b) applies a nonlinear activation function $\rho$ of type (ii).
Lastly, the network output is equal to the output of the map in question, as was established in the beginning of the proof.
This gives the result.
\end{proof}

With these pieces in place, let us construct a network that computes one iteration of Algorithm \ref{alg:NESTA}. Note that the $n$th iterate performs the update $z_n \mapsto z_{n+1}$. In order to write this as a neural network, we also need to keep track of the value of $q$ that is used to compute the intermediate vector $v_n$, since this value depends on not just $z_n$, but $z_0,\ldots,z_{n-1}$ as well. Hence, we now construct a network for the map
\begin{align}
    \begin{pmatrix} q_v^{(n-1)} \\ z_n \end{pmatrix}
    \mapsto
    \begin{pmatrix} q_v^{(n)} \\ z_{n+1} \end{pmatrix}. \label{eqn:unroll-NESTA-one-iteration}
\end{align}
Here $q_v^{(k)}$ refers to the value of $q$ used to calculate $v_k$ -- see lines 7--9 of Algorithm \ref{alg:NESTA}. 
We analogously define $q_x^{(k)}$, $\lambda_v^{(k)}$, $\lambda_x^{(k)}$ to be the values of $q$ and $\lambda$ used for $x_k$ and $v_k$, which are inferred from the notation. 
In the case of $n = 0$, we set $q_v^{(-1)} = z_0$, where $z_0$ is the initial point of NESTA.

\begin{lemma} \label{lem:unroll-NESTA-one-iteration}
The map (\ref{eqn:unroll-NESTA-one-iteration}) can be performed by a neural network $\cN \in \cN_{\bm{n},6,4}^*$ where
\begin{align*}
    \bm{n} = (2N, 2N+M, 2(N+m), 2(N+1), 3N+2, 3N+1, 2N)
\end{align*}
and the biases depend affinely on the measurement vector $y$ only. Moreover, the nonlinear activations are independent of $n$.
\end{lemma}

\begin{proof}
First let us write (\ref{eqn:unroll-NESTA-one-iteration}) as the sequence of maps 
\begin{align*} 
    \begin{pmatrix} q_v^{(n-1)} \\ z_n \end{pmatrix}
    \overset{T_1}{\longmapsto}
    \begin{pmatrix} q_v^{(n)} \\ q_x^{(n)} \end{pmatrix}
    \overset{T_2}{\longmapsto}
    \begin{pmatrix} q_v^{(n)} \\ q_x^{(n)} \\ \lambda_v^{(n)} \\ \lambda_x^{(n)} \end{pmatrix}
    \overset{T_3}{\longmapsto}
    \begin{pmatrix} q_v^{(n)} \\ z_{n+1} \end{pmatrix}.
\end{align*}
Regarding $T_1$, we know that the corresponding NESTA updates are
\begin{align}
    q_v^{(n)} &= q_v^{(n-1)} - \frac{\mu}{\beta} \alpha_n W \cT_\mu(W^* z_n), \label{eqn:q-v-update} \\
    q_x^{(n)} &= z_n - \frac{\mu}{\beta} W \cT_\mu (W^* z_n), \label{eqn:q-x-update}
\end{align}
for all $n \geq 0$. 
By Lemma \ref{lem:unroll-compute-q}, (\ref{eqn:q-v-update}) and (\ref{eqn:q-x-update}) can be expressed using a neural network $\cN_{x_n}', \cN_{v_n}' \in \cN_{(2N,N+M,N),2,1}^*$, where $\cN_{v_n}'$ uses $\alpha = \frac{\mu}{\beta} \alpha_n$ and $\cN_{x_n}'$ uses $\alpha = \frac{\mu}{\beta}$, so that
\begin{align*}
q_v^{(n)} = \cN_{v_n}' (q_v^{(n-1)}, z_n), \qquad q_x^{(n)} = \cN_{x_n}' (z_n, z_n).
\end{align*}
These networks can be run in parallel and be embedded into a larger network that models $T_1$. 
We do this by stacking the layers of $\cN_{v_n}'$ and $\cN_{x_n}'$ on top of each other and merge redundant copies of vectors and their network connections. 
Any missing connections simply correspond to zero weights. 
Note that a permutation of elements in the layers corresponds to a linear map for that layer before a nonlinear activation.
This procedure yields the map sequence (with affine and nonlinear activations combined into one map)
\begin{align*}
    \begin{pmatrix} q_v^{(n-1)} \\ z_n \end{pmatrix}
    \mapsto
    \begin{pmatrix} q_v^{(n-1)} \\ z_n \\ \cT_\mu (W^* z_n) \end{pmatrix}
    \mapsto
    \begin{pmatrix} q_v^{(n)} \\ q_x^{(n)} \end{pmatrix}
\end{align*}
Note that the only nonlinear activations used here are of type (i). 
The above map defines a network $\cN_{1} \in \cN_{(2N,2N+M,2N),2,1}^*$ that models $T_1$. 

For map $T_2$, by Lemma \ref{lem:unroll-compute-lambda} and the definition of Algorithm \ref{alg:NESTA}, $\lambda_v^{(n)}$ and $\lambda_x^{(n)}$ each can be expressed as the output of a network $\cN'' \in \cN_{(N,m,1,1),3,2}^*$, where $y$ in the lemma corresponds to the $y$ here.
Thus
\begin{align*}
    \lambda_v^{(n)} = \cN''(q_v^{(n)}), \qquad \lambda_x^{(n)} = \cN'' (q_x^{(n)}).
\end{align*}
Adopting the same strategy as we did for map $T_1$, we construct a network computing both $\lambda_v^{(n)}$ and $\lambda_x^{(n)}$ in parallel.
This gives the network layer sequence (with affine map and nonlinear activation combined per mapping)
\begin{align*}
    \begin{pmatrix} q_v^{(n)} \\ q_x^{(n)} \end{pmatrix}
    \mapsto
    \begin{pmatrix} q_v^{(n)} \\ q_x^{(n)} \\ \sigma_1(y - Aq_v^{(n)}) \\ \sigma_1(y - Aq_x^{(n)})\end{pmatrix}
    \mapsto
    \begin{pmatrix} q_v^{(n)} \\ q_x^{(n)} \\ \lambda_v^{(n)} \\ \lambda_x^{(n)} \end{pmatrix}
\end{align*}
where $\sigma_1$ is the squaring activation functiom from Lemma \ref{lem:unroll-compute-lambda}.
This map sequence embeds two copies of $\cN''$ and five identity maps. 
Note that we have implicitly included an identity map as the final affine map of the network.
By construction, and noting that there are only type (i) nonlinear activations here, $T_2$ is modelled by a network $\cN_2 \in \cN_{(2N,2(N+m),2(N+1),2(N+1)),3,2}^*$ with biases as affine maps of $y$.

Lastly, we construct a network that models $T_3$. 
Proceed by writing $T_3$ as the sequence of maps 
\begin{align*}
    \begin{pmatrix} q_v^{(n)} \\ q_x^{(n)} \\ \lambda_v^{(n)} \\ \lambda_x^{(n)} \end{pmatrix}
    \overset{\mathrm{(a)}}{\mapsto}
    \begin{pmatrix} 0 \\ \frac{\rho(\lambda_x^{(n)})}{c} A^* (y - Aq_x^{(n)}) \\ q_x^{(n)} \\ \lambda_v^{(n)} \\ q_v^{(n)} \end{pmatrix}
    \overset{\mathrm{(b)}}{\mapsto} 
    \begin{pmatrix} 0 \\ \frac{\rho(\lambda_v^{(n)})}{c} A^* (y - Aq_v^{(n)}) \\ x_n \\ q_v^{(n)} \end{pmatrix}
    \overset{\mathrm{(c)}}{\mapsto} 
    \begin{pmatrix} q_v^{(n)} \\ z_n \end{pmatrix}.
\end{align*}
Using Lemma \ref{lem:unroll-compute-x-v}, we have the following.
The map (a) applies nonlinear activation $\rho$ of type (ii) (from the proof Lemma \ref{lem:unroll-compute-x-v}) using scalar $\lambda_x^{(n)}$ after an affine mapping. 
Map (b) applies $\rho$ again with scalar $\lambda_v^{(n)}$ after an affine mapping. 
Lastly, the final map (c) is linear noting that $z_n = \tau_n v_n + (1 - \tau_n) x_n$.
Observe that the bias terms of the affine mappings are affine in $y$. 
This gives us the desired network corresponding to $T_3$, which is expressed as $\cN_3 \in \cN_{\bm{a},3,1}^*$ where
\begin{align*}
    \bm{a} = (2(N+1), 3N+2, 3N+1, 2N).
\end{align*}
Composing the networks to form $\cN = \cN_3 \circ \cN_2 \circ \cN_1 \in \cN_{\bm{n},6,4}^*$, noting that in the composition we merge each set of consecutively composed affine maps into a single affine map, gives a network that computes one iteration of NESTA (\ref{eqn:unroll-NESTA-one-iteration}).
Moreover $\cN$ has the property that all of its bias terms are affine maps of $y$ and the nonlinear activations do not depend on $n$. 
This completes the proof.
\end{proof}

Now we can unroll a network that computes $x_{n_{\max}}$ for $n_{\max} > 0$ of NESTA (Algorithm \ref{alg:NESTA}).
This is done by composing the network from Lemma \ref{lem:unroll-NESTA-one-iteration}.

\begin{theorem}[Unrolled NESTA] \label{thm:unroll-NESTA}
For $n \geq 0$, let $x_n$ be the $n$th iterate of NESTA (Algorithm \ref{alg:NESTA}) with input $y \in \bbC^m$ and initial point $z_0$. 
Then there exists a neural network $\cN \in \cN_{\bm{n},5(n+1)+1,4}^*$ with activation functions independent of $n$, and
\begin{align*}
    \bm{n} = (m, \underbrace{2N+M, 2(N+m), 2(N+1), 3N+2, 3N+1}_{\text{$n+1$ times}}, N),
\end{align*}
such that
\begin{align*}
    x_n = \cN(y).
\end{align*}
\end{theorem}

\begin{proof}
Consider the affine map $\phi : \bbC^m \rightarrow \bbC^{2N}$ defined by $u \mapsto (z_0, z_0)$ for $u \in \bbC^N$.
Let $\cN_0, \cN_{1}, \dots, \cN_{n-1}$ be copies of the network in Lemma \ref{lem:unroll-NESTA-one-iteration}.
Since we plan to compose $\cN_0$ with $\phi$, note that this defines $q_v^{(-1)} = z_0$.
Moreover, define $\cN_{n}$ as a modification of the network in Lemma \ref{lem:unroll-NESTA-one-iteration} by changing the linear map of its last layer to output $x_n$ instead of $(q_v^{(n)}, z_{n+1})$. This is straightforward to do by omitting $q_v^{(n)}$ and setting $\tau_n$ to be zero, since $z_{n+1} = \tau_n v_n + (1 - \tau_n) x_n$. 
Then the composition 
\begin{align*}
    \cN = \cN_{n} \circ \cN_{n-1} \circ \cdots \circ \cN_0 \circ \ \phi,
\end{align*}
where each set of consecutively composed affine maps are merged into a single affine map, gives the desired network in the theorem statement.
\end{proof}

Now we proceed with the main network construction we wanted, an unrolling of restarted NESTA.

\begin{theorem}[NESTANets: unrolled restarted NESTA] \label{thm:unroll-restarted-NESTA}
Let $x_{K+1}^\star$ be the output of restarted NESTA (Algorithm \ref{alg:restarted-NESTA}) with $K \geq 0$ restarts and $n_k = n \geq 0$ for all $k = 1, \dots, K+1$ corresponding to a fixed number of NESTA iterations for each restart.
Additionally, let $y \in \bbC^m$ be the input and $x_0^\star$ the initial point. 
Then there exists a neural network $\cN \in \cN_{\bm{n},5(K+1)(n+1)+1,4}^*$ with activation functions independent of $n$ and $K$, and
\begin{align*}
    \bm{n} = (m, \underbrace{2N+M, 2(N+m), 2(N+1), 3N+2, 3N+1}_{\text{$(K+1)(n+1)$ times}}, N),
\end{align*}
such that
\begin{align*}
    x_{K+1}^\star = \cN(y).
\end{align*}
\end{theorem}

Note the assumption that $n_k = n$ for all $k = 1, \dots, K+1$ is simply to leverage Theorem \ref{thm:NESTA-restart-error}, where the inner iterations $n_k$ are constant.

\begin{proof}
From the proof of Theorem \ref{thm:unroll-NESTA}, consider $\phi, \cN_0, \dots, \cN_{n-1}, \cN_{n}$.
Define $\cN_{n}'$ as a modification of $\cN_n$ that outputs $(x_n, x_n)$ instead of $x_n$.
Then defining the compositions
\begin{align*}
\cN_k^\star &= \cN_n' \circ \cN_{n-1} \circ \cdots \circ \cN_0, \quad k = 1, \dots, K, \\
\cN_{K+1}^\star &= \cN_n \circ \cN_{n-1} \circ \cdots \circ \cN_0,
\end{align*}
the desired network is 
\begin{align*}
    \cN = \cN_{K+1}^\star \circ \cN_{K}^\star \circ \cdots \circ \cN_1^\star \circ \phi.
\end{align*}
with each set of consecutively composed affine maps are merged into a single affine map.
\end{proof}

\section{Final arguments}\label{sec:final-args}

Now we prove the main result, Theorem \ref{thm:main-res}.

\begin{proof}
Fix $k \geq 1$ and $0 < r < 1$.
Let $c_1' = \frac{(1+\rho)^2}{1-\rho}$, $c_2' =  \frac{(3+\rho) \gamma}{1-\rho}$ be the constants $c_1$, $c_2$ stated in Theorem \ref{thm:NESTA-rNSP-error}, respectively.
Consider Algorithm \ref{alg:restarted-NESTA} defined with $K = k-1$ restarts and parameters specified in Theorem \ref{thm:NESTA-restart-error}. 
By Theorem \ref{thm:unroll-restarted-NESTA}, there exists a neural network $\cN \in \cN^*_{\bm{n},5k(n+1)+1,4}$ with
\bes{
n = \left\lceil \frac{2 c_1' \sqrt{\beta M}}{r \sqrt{s}}\right\rceil -1, \quad \bm{n} = (m, \underbrace{2N+M, 2(N+m), 2(N+1), 3N+2, 3N+1}_{\text{$k(n+1)$ times}}, N)
}
satisfying: for any input $y$ to the restarted NESTA algorithm, the final iterate $x_k^\star$ is computed by $\cN$, i.e. $\cN(y) = x_k^\star$.
Now we specify the input $y \in \bbM_{A,W,\chi,\eta}$ so that $y = Ax + e$ for some $(x,e) \in \bbI_{W,\chi,\eta}$.
Then using $\cN(y) = x_k^\star$ and Theorem \ref{thm:NESTA-rNSP-error}, we have
\bes{
\nmu{x - \cN(y)}_{\ell^2} \leq r^k \max\{\nmu{x}_{\ell^2}, \zeta\} + \frac{1-r^k}{1-r} \zeta.
}
By definition of the class $\bbM$ and $\zeta$, we have $\nmu{x}_{\ell^2} \leq 1$ and $\zeta \leq c \cdot \cC\cS_s(W^* x, \eta) \leq c \chi$, where $c = 2 \max\{c_1', c_2'\}$.
Using this, we further bound the error between $x$ and $\cN(y)$ to get
\bes{
\nmu{x - \cN(y)}_{\ell^2} \leq r^k \max\{1, c \chi\} + \frac{1-r^k}{1-r} c \chi \leq r^k + \left( r + \frac{1}{1-r} \right) c \chi.
}
Now referring to $c_1$ and $c_2$ as in the statement of the main result (Theorem \ref{thm:main-res}), these constants can be chosen as 
\bes{
{c_1 = \left(r+\frac{1}{1-r} \right)c, \qquad c_2 = \frac{5 c \sqrt{\beta}}{r} +6.}
}
Reading off the layer sizes in $\bm{n}$, the width of the network is at most $3N + M$.
Lastly, since $y$ was arbitrary, this completes the proof.
\end{proof}

\section{Numerical experiments} \label{sec:experiments}

We now present a series of numerical experiments, some of which reaffirm theoretical results and others that explore current gaps between theory and practice.
First, we demonstrate the exponential decay in the error bound of Theorem \ref{thm:main-res}.
Second, we compare the performance of NESTANets with and without restarts.
Third, we discuss and offer guidance on hyperparameter tuning.
Lastly, the fourth experiment demonstrates the stability of NESTANets by computing a worst-case perturbation of the measurements.

\subsection{Main example: Fourier imaging with the wavelets-plus-gradient transform}

Our main example pertains to Fourier imaging. Here $x \in \bbC^N$ is the vectorization of a complex, two-dimensional image $X \in \bbC^{n \times n}$, where $N = n^2$. The matrix $A$ takes the form
\bes{
A = \frac{1}{\sqrt{m}} P_{\Omega} F,
}
where $F \in \bbC^{N \times N} $ is the two-dimensional Fourier matrix, $\Omega \subseteq [N]$, $\abs{\Omega} = m$, is a sampling mask and $P_{\Omega} \in \bbC^{m \times N}$ is the \textit{row selector} matrix that selects the rows of $F$ corresponding to the indices in $\Omega$. Notice that
\be{
\label{A-A-star-prop}
A A^* = \frac{N}{m} I,
}
since $F F^* = N I$ and $P_{\Omega}$ is a selector matrix. Hence, this matrix satisfies the property required in Theorem \ref{thm:main-res} with $\nu = N/m$.

For analysis-sparsity operator $W$ we consider the concatenation of the discrete gradient and Haar wavelet operators.
Specifically,
\be{
\label{Haar-gradient-analysis}
W^* = \begin{bmatrix} \Phi^* \\ \sqrt{\lambda} \nabla \end{bmatrix} \in \bbR^{3 N \times N},
}
where $\Phi \in \bbR^{N \times N}$ is the orthogonal, two-dimensional discrete Haar wavelet basis (see, e.g., \cite[Sec.\ SM2.1]{adcock2021improved}) and $\nabla \in \bbR^{2 N \times N}$ is the two-dimensional (anisotropic) discrete gradient operator with periodic boundary conditions (see, e.g., \cite[Sec.\ 2.5]{adcock2021improved}). Here $\lambda \geq 0$ is a parameter that controls the relative influence of each sparsifying transform. This analysis operator is popular in compressive imaging -- especially medical imaging applications, including commercial MR scanners \cite{fessler2020optimization} -- since it combines the ability of the discrete gradient operator to produce sharp edges with the wavelet transform to capture smooth variations \cite{fessler2020optimization,lustig2008compressed,lustig2007sparse}. However, we do not claim that it is the `best' sparsifying transform available. Other transforms such as curvelets \cite{candes2000curvelets---a,candes2002recovering,candes2004tight}, ridgelets \cite{candes1999ridgelets:} or shearlets \cite{labate2005sparse,guo2006sparse,guo2007optimally,kutyniok2011compactly} may yield superior performance in certain imaging scenarios.

Notice that the analysis operator \R{Haar-gradient-analysis} satisfies Assumption \ref{ass:W-frame}. Indeed, since $\Phi$ is orthogonal we have
\bes{
\nm{x}^2_{\ell^2} \leq \nm{x}^2_{\ell^2} + \lambda \nm{\nabla x}^2_{\ell^2} = \nm{W^* x}^2_{\ell^2} \leq (1+ 8 \lambda) \nm{x}^2_{\ell^2}.
}
Here we used the fact that $\nm{\nabla}^2_{\ell^2} \leq 8$. 
Hence Assumption \ref{ass:W-frame} holds with upper frame bound $\beta \leq 1 + 8 \lambda$.

\begin{remark}

It is possible to construct random sampling masks such that the pair $(A,W)$ satisfies the rNSP with high probability, provided $m \gtrsim s \cdot \mathrm{polylog}(s,N)$. We now sketch such a construction. 

The basic idea is to construct $\Omega = \Omega_1 \cup \Omega_2$, where $\Omega_1$ and $\Omega_2$ are chosen so that the pairs $(A_1,\Phi)$ and $(A_2,\nabla^*)$ satisfy the rNSP with high probability, where $A_i = \frac{1}{\sqrt{m}} P_{\Omega_i} F$, $i = 1,2$. For the first pair, this can be achieved by choosing the frequencies in $\Omega_1$ randomly and independently according to an \textit{inverse square law} density, wherein a frequency $\omega = (\omega_1,\omega_2)$ is included in $\Omega_1$ with probability proportional to $1/(\omega^2_1 + \omega^2_2)$. It is well-known that such a matrix, after a diagonal scaling, satisfies the Restricted Isometry Property (RIP) with high probability (see, e.g., \cite{krahmer2013stable} or \cite[Lem.\ 13.20]{adcock2021compressive}), and therefore the rNSP. For the second pair, this can be achieved by choosing the frequencies in $\Omega_2$ randomly and independently according to the uniform density, and then by making use of the \textit{commuting property} of the discrete gradient operator and the Fourier matrix (see, e.g., \cite{poon2015role} or \cite[Sec.\ 17.4.1]{adcock2021compressive}). Since both pairs satisfy the rNSP, it follows that the pair $(A,W)$ of concatenated matrices also does.

The problem with this construction is that the matrix $A$ does not satisfy \R{A-A-star-prop}. The reason is that the sampling masks $\Omega_i$ may contain repeats, since frequencies are chosen randomly with replacement (this is done in order to ensure independence of the rows of $A_i$, which facilitates proving the RIP). Theoretically, this can be overcome by using a \textit{Bernoulli sampling model}, wherein each frequency is selected based on the outcome of a biased coin toss (see, e.g., \cite[Sec.\ 11.4.3]{adcock2021compressive}). First, one generates $\Omega_1$ according to a Bernoulli model with probabilities chosen according to an inverse square law. Then, conditional on $\Omega_1$, one chooses $\Omega_2 \subseteq [N] \backslash \Omega_1$ according to a Bernoulli model with equal probabilities. This model ensures the rNSP for the pair $(A,W)$ and, since each frequency is selected at most once, the property \R{A-A-star-prop}. The downside is that the number of measurements is now itself a random variable, equal to $m$ in expectation only. Despite this, Bernstein inequalities (or Chernoff bounds) can show that the true number of measurements in a Bernoulli model sampling scheme does not deviate far from $m$.

Since the full construction of this sampling scheme is rather technical and not a primary focus of this paper, we omit further details (including a proof that the pair $(A,W)$ satisfies the rNSP). These will be presented in an upcoming paper.
\end{remark}

\subsection{Setup}

We implement NESTANets, i.e. NESTA and restarted NESTA (Algorithm \ref{alg:NESTA} and \ref{alg:restarted-NESTA}) to solve the Fourier imaging problem introduced in the previous section.
The implementation is written in Python using PyTorch \cite{paszke2019pytorch}, an open-source machine learning software that offers a wide scope of tools to implement neural networks and manipulate arrays.
PyTorch also supports GPU acceleration and automatic differentiation, both of which are crucial to running our stability experiments of NESTANets at a large scale.
The remaining experiments can easily be run without a GPU on a standard desktop computer.
The code for the experiments shown in this section can be found in 

\begin{quote}
\centering
\url{https://github.com/mneyrane/AS-NESTA-net}.
\end{quote}

\begin{figure}
\centering
\includegraphics[width=128px]{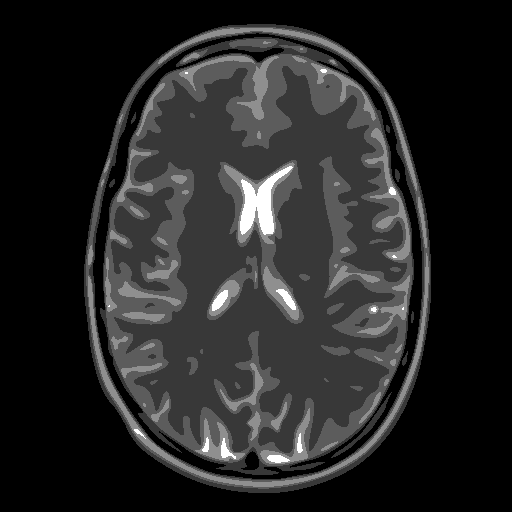}
~
\includegraphics[width=128px]{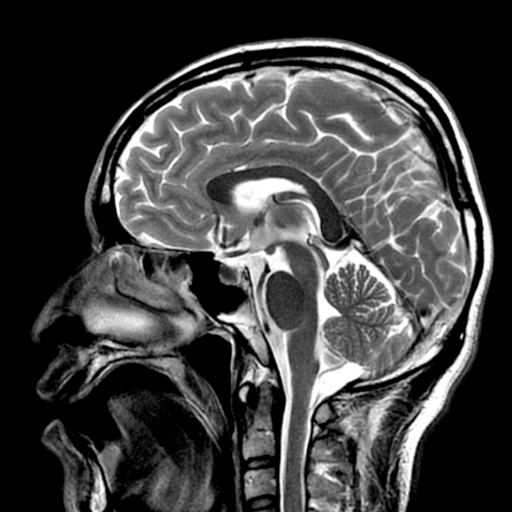}
\caption{Test images, GPLU phantom (left) and a brain MR image (right).}
\label{fig:test-images}
\end{figure}

A key aspect of designing the experiments is choosing hyperparameters, for which there are several.
The first set define the problem to solve, i.e. the wavelet-gradient weight $\lambda \geq 0$ and the sampling rate $m/N$.
The second set defines the NESTA solver, namely the number of restarts $K$, decay factor $r$, and special constants $\eta$, $\zeta$, and $\delta$. 
The number $\eta$ defines the noise level for the solver. 
We define $\delta = \frac{\sqrt{s}}{c_1 M}$ from Theorem \ref{thm:NESTA-restart-error} (and consequently $0 < \delta < 1/\sqrt{M}$) and, slightly abusing notation, $\zeta$ is any upper bound to the $\zeta$ value appearing in Theorem \ref{thm:NESTA-rNSP-error}.
Thus, $\delta$ determines the number of inner iterations $n = \lceil (2 \sqrt{\beta})/(r \delta \sqrt{M}) \rceil - 1$ and $\zeta$ represents the target error level.
Combined with $\zeta$, $\delta$ also determines the smoothing parameters $\{\mu_k\}$.
The parameter $\eta$ defines the constraint in \eqref{eqn:W-QCBP} for the solver.
Unless stated otherwise, we fix $\lambda = 5/2$, $r = 1/4$, and $\zeta = 10^{-9}$.
The remaining parameters are defined per experiment.
We also use two test images shown in Figure \ref{fig:test-images}.
The GPLU phantom image \cite{guerquin2012realistic} is suitable for testing Fourier imaging techniques, since it is a piecewise constant image.
In particular, the phantom image is very close to being exactly sparse under the wavelet-gradient transform.
In the cases where we calculate relative errors, note that the (vectorized) phantom image has an $\ell^2$-norm of $\approx 122.93$.
We use it for all but the stability experiment, where we use the brain MR image (Fig.\ \ref{fig:test-images}, right).

\subsection{Results and analysis}

\subsubsection{Exponential decay of reconstruction error}

\begin{figure}
\centering
\includegraphics[width=0.65\textwidth]{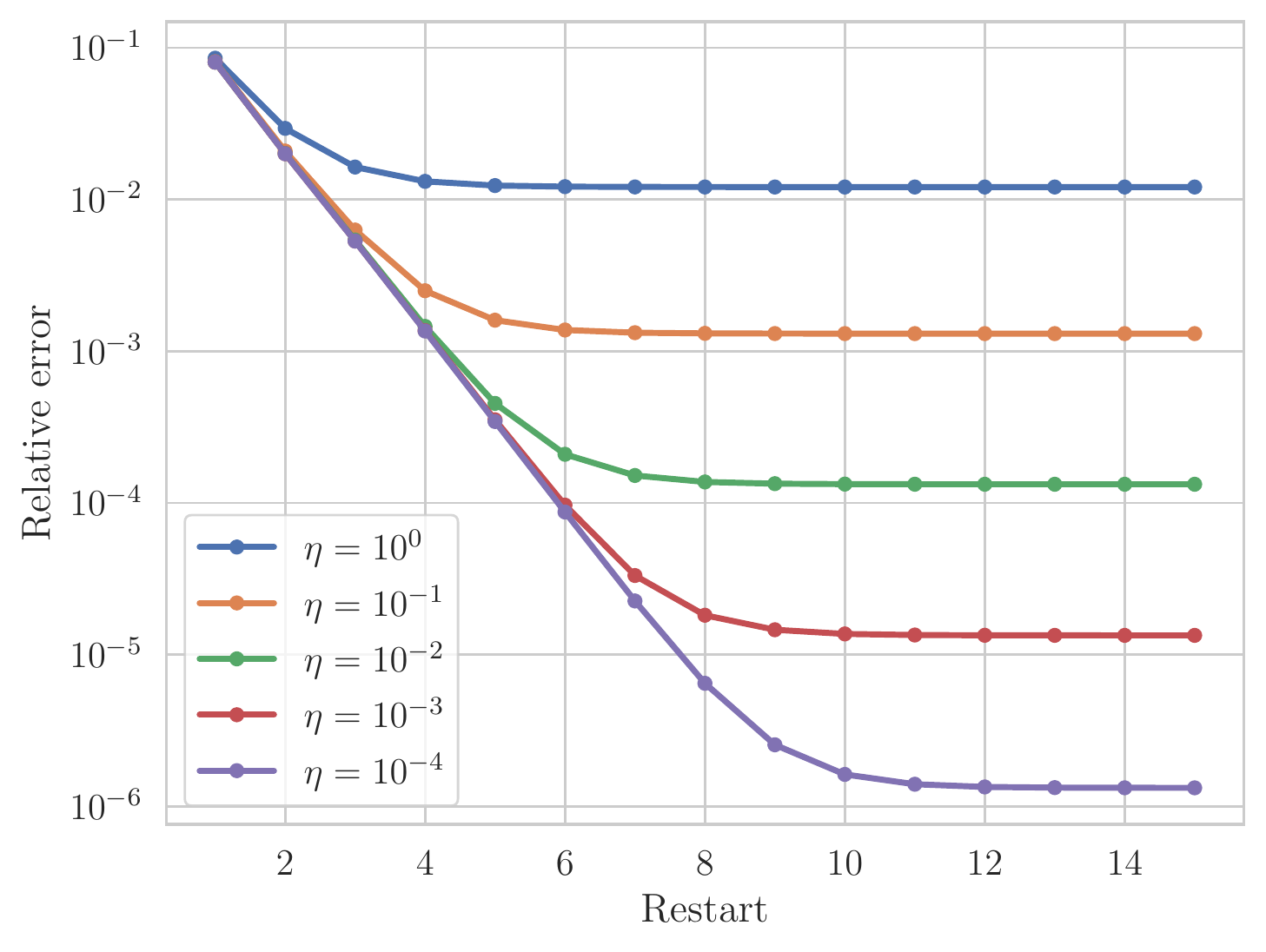}
\\
\includegraphics[width=0.65\textwidth]{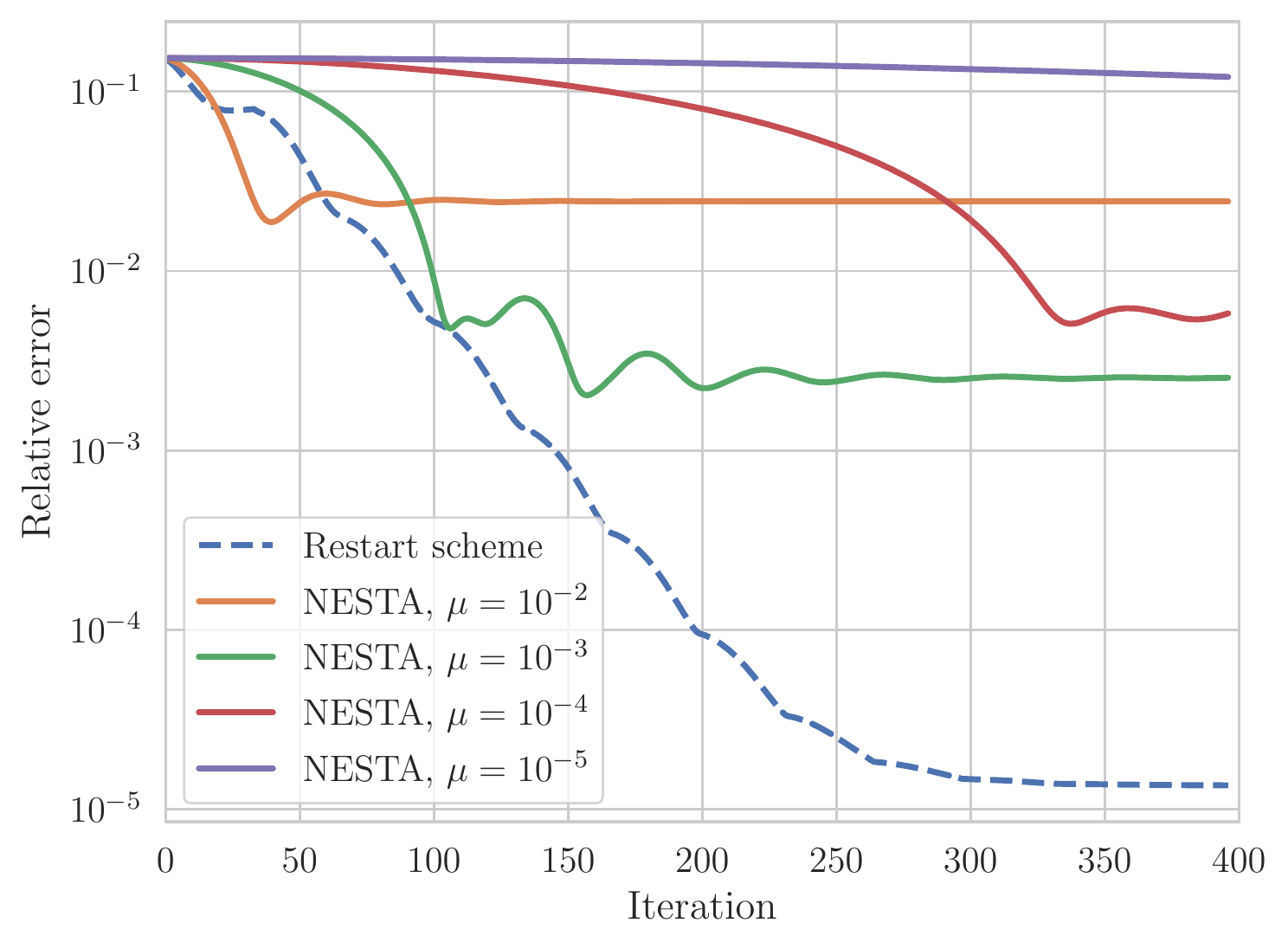}
\caption{
Top plot shows relative error $\nmu{x_k^\star - x}_{\ell^2}/\nmu{x}_{\ell^2}$ versus $k$ for different values of $\eta$, displaying exponential decay.
Bottom plot shows relative error versus total inner iteration for NESTA with and without restarts. The smoothing parameter value $\mu$ is varied for the no-restart case.
}
\label{fig:exp-decay-AND-without-res}
\end{figure}

Here we fix a 15\% sampling rate, $K = 14$ restarts, $\delta = 1.25 \cdot 10^{-3}$, and $\eta = 10^{-i}$ for fixed $i = 0, 1, 2, 3, 4$.
Each restart iteration runs 33 inner iterations.
The measurements are computed by $y = Ax + e$ where $e$ is random Gaussian noise rescaled to have $\nm{e}_{\ell^2} = \eta$. 
The relative restart iterate error $\nmu{x_{k}^\star - x}_{\ell^2} / \nmu{x}_{\ell^2}$ for each $k$ is plotted in Figure \ref{fig:exp-decay-AND-without-res}.
In each case of $\eta$, the error decays exponentially to a limiting error of approximately $\eta / \nmu{x}_{\ell^2}$.
This is consistent with Theorems \ref{thm:main-res} and \ref{thm:NESTA-restart-error}.
The GPLU phantom is piecewise constant, and is thus approximately sparse under the Haar wavelet and discrete gradient transforms, so $\sigma_s(W^* x)_{\ell^1} \approx 0$. 
As we expect, the error is mostly from the noise level $\eta$.

\subsubsection{Comparing NESTA with and without restarts}

To compare NESTA with and without a restart scheme, we reuse the same parameters from the previous experiment, except $\eta = 10^{-3}$, $K = 11$. 
In the no-restart case, we fix smoothing parameter $\mu = 10^{-i}$, $i = 2, 3, 4, 5$.
Both with and without restarts, NESTA computes 396 iterates.
Figure \ref{fig:exp-decay-AND-without-res} plots the relative iterate error $\nmu{x_t - x}_{\ell^2} / \nmu{x}_{\ell^2}$ for each iterate $t$.
Similar to the previous experiment, the restart scheme exhibits exponential decay to a limiting error.
No-restart NESTA for larger values of $\mu$ converges quickly but to a larger limiting error than that which the restart scheme produces.
Using smaller $\mu$ leads to a lower error, but at the cost of requiring more iterations to get a lower error.
This is consistent with what is observed in theory, namely the error bound established in Theorem \ref{thm:NESTA-rNSP-error}. As expected, the restart scheme excels in performance in the low-noise regime.

In contrast to no restarts, the restart scheme controls the smoothing parameters using the values $\zeta$, $\delta$ and $r$. 
The parameter $\zeta$ (and also $\eta$) control the extent of the limiting error.
Here $r$ controls the rate of convergence and $\delta$ influences the precision of the recovery. We elaborate more on why this is in the hyperparameter tuning discussion coming next.

\subsubsection{Tuning hyperparameters}

For this section, we discuss how to select the restarted NESTA parameters $K$, $r$, $\zeta$, $\eta$ and $\delta$.
Note that the experiment presented here is specific to the tuning of $\zeta$ and $\eta$, where as the rest correspond to general guidelines and observations.

We select our hyperparameters around Theorem \ref{thm:NESTA-restart-error}.
A consequence of this approach is a gap between theory and practice. 
In practice, it is hard to know the true values of some of these parameters a priori, since some depend on difficult-to-compute compressed sensing constants, and one generally lacks access to the true signal being recovered.

For $K$, a simple approach to choose an ``optimal'' $K$ is to compute one from Theorem \ref{thm:NESTA-restart-error}, for fixed $r$, $\zeta$, and precision $\alpha$ to get an error bound of $\alpha + \zeta/(1-r)$.
Otherwise, choosing $K$ is a matter of trial and error.

The parameter $r$ controls the rate of convergence and limiting error bound, but is also inversely proportional to $n_k$. 
This tells us that for faster convergence per restart and a lower error, we need more inner iterations.
An optimal choice of $r$ can be derived when we fix a budget of \textit{total iterations}, and then minimize the decaying error term involving $r^k$. This can be done by expressing the number of restarts in terms of $r$ using Theorem \ref{thm:NESTA-restart-error}. Our implementation of restarted NESTA does not treat total iterations as a parameter, so we simply resort to manually selecting $r$.

\begin{figure}
\centering
\includegraphics[width=0.65\textwidth]{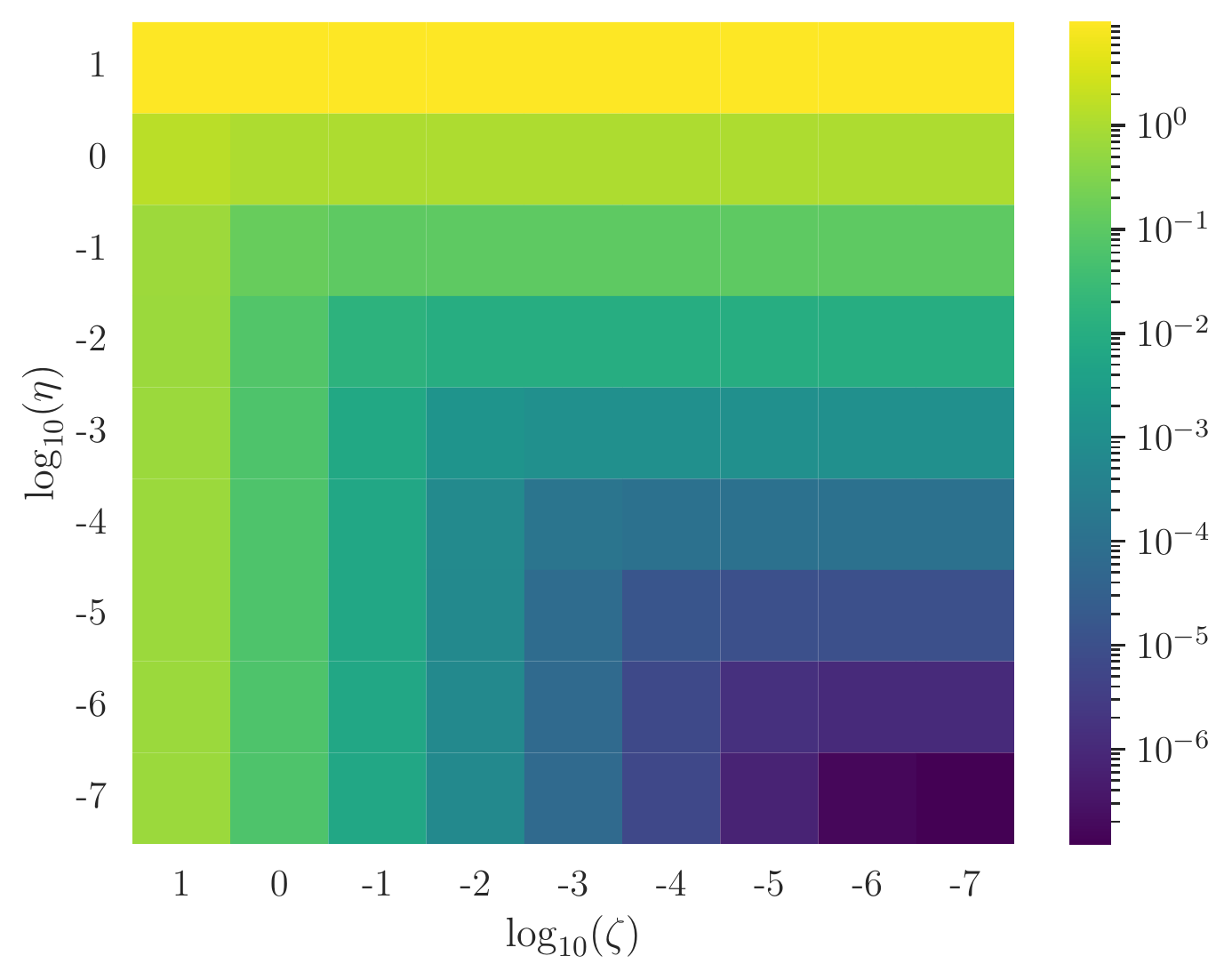}
\caption{Contours of the error $\nmu{\hat{x} - x}_{\ell^2}$ where $\hat{x}$ is the final iterate of restarted NESTA, for different values of $\eta$ and $\zeta$.}
\label{fig:error-contours}
\end{figure}

To address the choice of $\zeta$, what we found empirically is that it suffices to pick $\zeta < \eta$. 
To show this, we run an experiment that uses the same parameters and image from the exponential decay experiment, except that the sampling rate is now 25\%, $y = Ax$ and $\eta, \zeta = 10^{-i}$ for $i = -1, 0, 1, \dots, 7$.
The higher sampling rate is chosen to ensure a high precision reconstruction for lower noise levels.
In Figure \ref{fig:error-contours}, we plot the reconstruction error of the final restarted NESTA iterate for different values of $\eta$ and $\zeta$.
The contours are approximately of the form $\max\{\eta, \zeta/10 \}$, which intuitively suggests that restarted NESTA cannot produce a reconstruction much better than the assumed noise level $\eta$ or error level $\zeta$.
This suggests that one can choose $\zeta$ to be less than the true error value without sacrificing accuracy.
To contrast with the theory, Theorem \ref{thm:NESTA-restart-error} assumes $\zeta$ is an upper bound for the error, which is then used to prove the exponential decay in the restart scheme's reconstruction error. However, this experiment suggests that we do not need to treat $\zeta$ as an upper bound in practice. As stated earlier, $\zeta$ can be as small as possible, or even zero, provided the computed smoothing parameters $\{\mu_k\}$ do not fall below machine precision.

Regarding the choice of $\delta$, there are a few comments to make.
By definition, $\delta$ is proportional to $\mu_k$ and inversely proportional to $n_k$, suggesting a trade-off between quality of reconstruction and number of algorithm iterations.
The underlying difficulty is determining an \textit{optimal} choice of $\delta$. 
Having an optimal value of $\delta$ is desirable to avoid being short of the best reconstruction and expending unnecessary calculations of iterates.
As noted before, a priori information to select $\delta$ is absent in a practical setting, so the best one can do to guarantee a good quality reconstruction is to set $\delta$ as small as reasonably possible.
In particular, $\delta$ needs to be much smaller to obtain good reconstructions at lower sampling rates.

\subsubsection{Stability}

To test stability, given measurements $y = Ax$ we strive to compute a worst-case perturbation \cite{antun2020instabilities} $e$ of the measurements $y$ that maximizes the difference between $\cN(y)$ and $\cN(y+e)$, where $\cN : \bbC^{m} \rightarrow \bbC^N$ is a NESTANet.
More precisely, we solve
\be{
\label{eq:worst-pert}
\max_{e \in \bbC^{m}} \nmu{\cN(y) - \cN(y+e)}_{\ell^2}^2 \quad \mbox{subject to $\nmu{e}_{\ell^2} \leq \tilde{\eta}$} 
}
similar to \cite{genzel2022solving}.
Here $\tilde{\eta}$ is a parameter controlling the maximum size of the perturbation.
To solve \eqref{eq:worst-pert}, we perform projected gradient ascent for a fixed number of iterations, noting the projection onto the feasible set of \R{eq:worst-pert} is straightforward to compute.
The initial value of $e$ is chosen randomly over an $\ell^2$-ball of radius $\tilde{\eta} / \sqrt{m}$, and over all the gradient ascent iterates we select the one that produced the largest objective value of \eqref{eq:worst-pert}.

Solving \R{eq:worst-pert} exactly allows one to determine the local $\tilde{\eta}$-Lipschitz constant of the reconstruction map $\cN$, and therefore, if small, assert stability of $\cN$ at a given $y$.
However, \eqref{eq:worst-pert} is a high-dimensional nonconvex optimization problem, so we can at best approximate local optima. Also, finding local optima is now sensitive to the choice of step size of gradient ascent and the random initialization of $e$. For these reasons, numerical results indicative of stability fall short of being a conclusive verification of it.

\begin{figure}
\centering
\includegraphics[width=0.8425\textwidth]{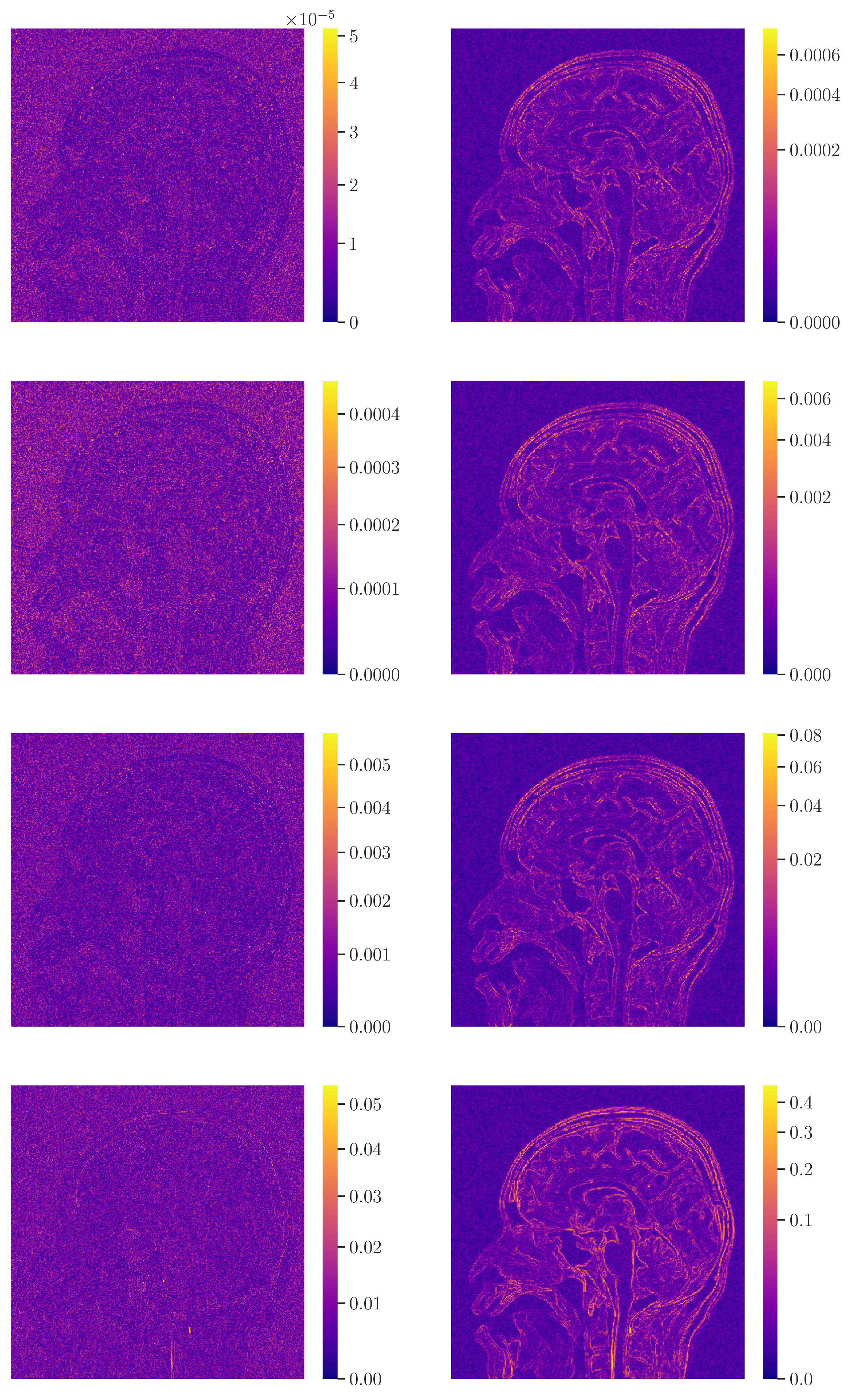}
\caption{Colour plots of worst-case perturbation $e$ in the image domain via the psuedoinverse of $A$, i.e. $\lvert A^\dagger e \rvert$ (left column), and the reconstruction difference $\lvert \cN(y + e) - \cN(y) \rvert$ (right column). The absolute value is applied elementwise. Each row from top to bottom corresponds to the noise level $\tilde{\eta} = 10^{i} \eta$ for $i = 0,1,2,3$ of the computed worst-case perturbation $e$. To help visualization, the plots in the left and right column use a power-law colourmap rescaling of $4/5$ and $2/5$, respectively.}
\label{fig:stability}
\end{figure}

Since automatic differentiation is used to compute the derivative of the objective in \eqref{eq:worst-pert}, this optimization is computationally expensive when many restarts or inner iterations are used.
This limits running this experiment on larger unrolled networks. Hence in this experiment we lower the number of unrolled iterations. Specifically, we use a 25\% sampling rate, $K = 9$ restarts, $\delta = 2.33 \cdot 10^{-3}$, $\eta = 10^{-2}$ and $\tilde{\eta} = 10^{i} \eta$, with $i = 0, 1, 2, 3$.
Each restart iteration computes 17 inner iterations.
We construct worst-case perturbations $e$ to measurements $y = Ax$ where $x$ is the brain MR image in Figure \ref{fig:test-images}.
We perform 400 trials of gradient ascent for each value of $\tilde{\eta}$, each consisting of 150 ascent iterations with a step size of $3.0$.
For each value of $\tilde{\eta}$, we plot the worst-case perturbation maximizing \eqref{eq:worst-pert} and the reconstruction of the perturbed measurements, in Figure \ref{fig:stability}.
The perturbations are represented in the image domain by applying the Moore-Penrose pseudoinverse $A^\dagger = \nu^{-1} A^*$ to the perturbations.
The use of colour plots are to help visualize the perturbation and reconstruction, moreso for lower noise levels where visual artefacts are hard (or impossible) to see in the grayscale image.

\begin{figure}
\centering
\includegraphics[width=75px]{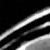}
\hfill
\includegraphics[width=75px]{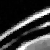}
\hfill
\includegraphics[width=75px]{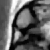}
\hfill
\includegraphics[width=75px]{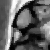}
\caption{Crops of $\cN(y)$ (left in each crop) and $\cN(y+e)$ (right in each crop) for the computed worst-case perturbation $e$ with $\tilde{\eta} = 1000 \eta$. These are a grayscale rendering of the clipped elementwise absolute value of the reconstructions.}
\label{fig:large-pert-fx}
\end{figure}

What we observe for $\tilde{\eta} = \eta$ is the stability we expect from the theory, e.g. Theorem \ref{thm:NESTA-restart-error}.
In fact, even for larger perturbations beyond the assumed noise level ($\tilde{\eta} > \eta$), the algorithm remains stable.
It is possible to theoretically justify this observation, see, e.g. \cite{adcock2018robustness}, at least in the no-restarts case. This experiment suggests that NESTANets are also stable to perturbations exceeding the assumed noise level. Interestingly, the worst-case perturbations for each $\tilde{\eta}$ tend to increase reconstruction error near the discontinuities in the image, i.e. boundaries of piecewise smooth components.
A closeup of this is shown in Figure \ref{fig:large-pert-fx} for $\tilde{\eta} = 1000\eta$, where minor visual artefacts appear along edges in the reconstruction of the perturbed measurements.

\section{Conclusions}\label{sec:conclusion}

The purpose of this work is to demonstrate that stable deep neural networks can be constructed for inverse problems where the underlying objects are approximately analysis-sparse in an arbitrary frame. Our neural network construction NESTANets relies on a novel unrolling of NESTA, in combination with a restart procedure, which yields deep neural networks whose depth scales only logarithmically with the desired accuracy. In doing so, we add further mathematical credence to the great potential of deep neural networks and deep learning for solving inverse problems.

There are a number of avenues for further research. First, on the theoretical side, is extending these results to instances where the analysis operator does not form a frame. For example, we can consider approximately gradient-sparse signals. Gradient sparsity can be addressed through a more sophisticated analysis, which we intend to present in a subsequent paper. Second, there are other popular redundant analysis operators such as curvelets \cite{candes2000curvelets---a,candes2002recovering,candes2004tight}, ridgelets \cite{candes1999ridgelets:} or shearlets \cite{labate2005sparse,guo2006sparse,guo2007optimally,kutyniok2011compactly}. Although these do form frames, it is less well known how to design (Fourier) sampling strategies for these systems that satisfy the rNSP. See \cite{kutyniok2018optimal} for some work in this direction. Third, it is known that the analysis sparsity model presented in Section \ref{ss:main-res} has some limitations. As has been discussed in \cite{nam2013cosparse,genzel2021analysis}, for redundant dictionaries this may suggest substantially more number of measurements than needed in practice. More refined analyses are therefore an interesting topic for future work.

Another line of extension is to address whether smoothing, in general, is a good approach to obtain neural networks from unrolling. To the best of our knowledge, smoothing has not previously been used as part of unrolling schemes for deep neural network constructions, while other standard techniques for dealing with nonsmooth objective functions (e.g., splitting, proximal maps, etc.) are frequently used.

Finally, perhaps the most significant question pertains to the practical impact of unrolling. Unrolling is used in some of the more effective deep learning procedures for inverse problems in imaging. However, as noted, this does not ensure stability and an absence of hallucinations. How to combine unrolling with its mathematical underpinnings in compressed sensing theory, together with learning to outperform state-of-the-art model-based methods while maintaining stability, is a significant open problem.

\backmatter

\bmhead{Acknowledgments}

BA acknowledges support from NSERC through grant R611675. MNN acknowledges support from an NSERC CGS-M scholarship. Both authors would like to thank Vegard Antun and Matthew Colbrook for helpful advice and feedback. On behalf of all authors, the corresponding author states that there is no conflict of interest.

\bibliography{refs}

\end{document}